\documentclass[lettersize,journal]{IEEEtran}
\usepackage{amsmath,amsfonts}
\usepackage{algorithmic}
\usepackage{array}
\usepackage[caption=false,font=normalsize,labelfont=sf,textfont=sf]{subfig}
\usepackage{textcomp}
\usepackage{stfloats}
\usepackage{verbatim}
\usepackage{graphicx}
\def\BibTeX{{\rm B\kern-.05em{\sc i\kern-.025em b}\kern-.08em
    T\kern-.1667em\lower.7ex\hbox{E}\kern-.125emX}}
\PassOptionsToPackage{hyphens}{url}\usepackage{hyperref}
\usepackage{url}
\usepackage{balance}
\usepackage{xcolor}
\begin{document}
\title{Cellular Predictions on the Move: What about Data?}
\author{Natalia Vesselinova and Pauliina Ilmonen 
\thanks{The authors gratefully acknowledge the support received from Academy of Finland via the Centre of Excellence in Randomness and Structures, decision number 346308 and the computational resources provided by the Aalto Science-IT project.}}

\markboth{Journal of \LaTeX\ Class Files,~Vol.~18, No.~9, September~2020}%
{How to Use the IEEEtran \LaTeX \ Templates}

\maketitle

\begin{abstract}
Mobile cellular load forecasting is native to network 
resource optimization and delivery of services 
with reliability, latency and quality guarantees. 
The mainstream of  
machine learning research 
in the area
is focused primarily on 
developing powerful
learning structures
for improved prediction accuracy.
The data used 
for forecasting
traditionally belong to
the cellular domain
and at most 
contain exogenous information 
about the surroundings
of the  base stations.
We approach the prediction task
from the perspective of data
as a vital component of any 
data learning process. 
We hypothesize that 
substantial improvements 
could be achieved 
when the data inform on
the processes that 
create the cellular load. 
Specifically, 
we propose to characterize 
the population dynamics---the
potential number of 
cellular traffic sources 
and their mobility---in
addition to employing 
historical time series 
of mobile data traffic.
We validate our hypothesis 
for the rarely 
examined highway scenario.
Comprehensive experiments 
show forecasting improvements
on the order of $60\%$
due to the use of
these data alone.
\end{abstract}


\section{Introduction}


\IEEEPARstart{C}{onnected} and automated 
mobility services
are aimed at 
supporting 
autonomous vehicles 
and cooperative driving
as part of greener 
transportation systems.
Such solutions 
are developed
with the intent to 
reduce traffic accidents
and congestion, and
the associated 
harmful emissions
and costs.
They are designed to 
enhance mobility 
by ensuring 
smooth, efficient and
secure movement 
within and across 
national borders.
Highways 
are identified 
as major terrestrial transport paths
for national and  international mobility
of people, commerce, 
and freight transport. 
Therefore, they have received
special attention 
on the path to 
seamless connectivity. 
Specifically, several large 
European Union
(EU) 
initiatives
have funded 
5G trials
of connected 
and automated
mobility services 
on highways,  
 the provisioning 
of which is aligned 
with major 
EU
sustainability goals.

Mobile cellular load forecasting 
is native to network resource 
optimization and delivery 
of services with 
quality guarantees. 
In effect, 
connected and 
automated mobility 
is supported by
safety-critical applications, 
which demand 
very low latency 
and high reliability.
Cellular load predictions
are vital for
guaranteeing 
these demands.
Both load prediction
and 
resource reservation 
occur at three different 
time scales \cite{Bega20}. 
Long-term and mid-term  
forecasting assist 
the management 
and orchestration 
domain in deciding 
on the placement
of (virtual) resources
and on resource
scaling
at a coarse granularity
(from hours to
 tens of minutes, 
respectively).
On the other hand,
short-term forecasting 
supports the 
orchestration of  
radio access network
resources---optimization
and allocation 
of resources---at
fast time scales
(on the order of
 seconds to 
a couple of minutes).
Forecasting gains 
further relevance 
 on highways,
 where vehicular speeds
 are the highest and
 flow fluctuations
are the fastest.
This motivates 
our focus on
short-term 
resource forecasting 
on highways.

In recent years, 
the primary focus of 
the mainstream research 
in the area of
mobile cellular 
forecasting 
has been on the design
of advanced 
machine learning structures.
Few forecasting studies 
examine the role 
and usefulness of data.
Other than cellular network 
key performance indicators (KPIs), 
the main interest 
has been in 
additional information 
that can describe
the environment 
within which 
a base station (BS) operates.
Typically, the BS context 
is characterized by 
the distribution of 
points of interest (PoI). 
The latest research 
in the area introduces 
knowledge graphs (KGs). 
In their essence, 
KGs are an extension to
the PoI, where 
the environmental information
about the BS surroundings 
is enriched further.

Contrastingly, we propose 
to learn the intrinsic forces 
that govern the 
mobile cellular traffic
generation.
We hypothesize that this
will scale up significantly
the forecasting accuracy. 
Specifically,
in addition to 
cellular time series,
we propose 
to employ  
population dynamics data.
We model 
the potential sources 
of cellular load
and their mobility
on highways
with vehicular flow
and speed.
In summary, our contributions are:
\begin{itemize}
	\item To the best of our knowledge,
				this is the first approach  
				designed to capture 
				the variables inherent to
				the mobile traffic generation process. 
				We employ vehicular metrics
				to model population dynamics 
				on highways and roads.
	\item We develop a methodology to 
				generate mobile cellular data
				based on road measurements,
				which overcomes the limitations
				imposed by mobile data privacy concerns.
	\item We extensively evaluate
				and validate our hypothesis for
				an existing highway featuring 
				highly fluctuating road traffic and
				mobile load, as well as seasonal changes.
	\item We examine highways and address short-term mobile
	cellular traffic predictions in contrast to the extensively
	studied urban scenario and long-term  forecasting.
				
\end{itemize}

\begin{figure*}[htbp]
	\centerline{\includegraphics[width=0.98\textwidth]{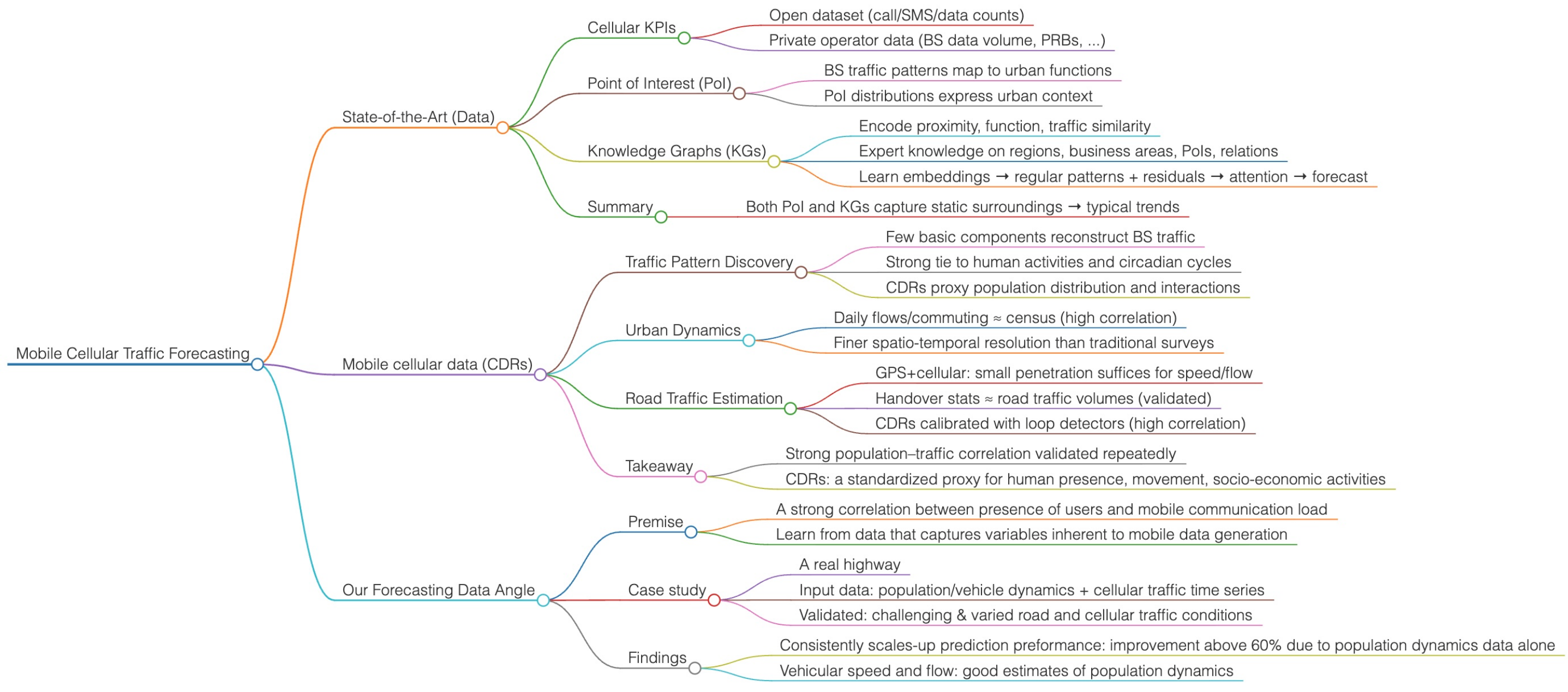}} 
	\caption{Schematic overview of the concept: background, motivation and our approach. Prior art employs pure cellular data, or at most external factors that shape the mobile data profile. Human activitivities are imprinted in the use of the cellular networks---there is a strong correlation between the presence of users and mobile data traffic. We propose to learn from data that informs the model on the variables that create the load on the mobile system.}
	\label{fig:sota}
\end{figure*}

This article builds upon 
our previous work. 
In \cite{vesselinova2023road}, 
we introduce 
the idea of employing data 
that inform on the sources
of radio access load
and their mobility
for improved forecasting 
of cellular traffic.
As a proof-of-concept 
study \cite{vesselinova2023road},
the idea is evaluated  
under controlled 
conditions---specific 
call intensities, 
handover rates and 
radio range.
Further, 
the concept is verified
in \cite{vesselinova2025data} 
under diverse road  
and cellular conditions
with very few
control parameters
as our goal is to mimic
real cellular
conditions.
The application of our concept
for supporting 
adaptive network slicing in 5G
is described in \cite{NV}.
In this study,
which addresses
the short-term 
forecasting problem 
on highways
(Section~\ref{sec:prediction})
too,
we review
the existing 
state-of-the-art
from the perspective 
of data, 
survey research 
that employs mobile cellular records
for modeling population dynamics,
elaborate on the concept 
and the data used to validate
our approach
(Section~\ref{sec:what_about_data}).
We reinforce our 
earlier results by
investigating the concept 
under different 
probability distributions
(Section~\ref{sec:experiments}),
 examining 
the  effect of 
the distribution on 
the prediction accuracy.
This study 
also brings 
new and deeper 
insights into the 
obtained results and 
observed trends 
(Section~\ref{sec:results}),
emphasizes the impact
of our approach
(Section~\ref{sec:discuss}),
and concludes with future prospects
(Section~\ref{sec:conclusion}).
Figure~\ref{fig:sota}
presents 
the conceptual framework 
of the study.

\section{Cellular Predictions: Problem Formulation}
\label{sec:prediction}

Cellular traffic is characterized by
temporal variations 
and spatial correlations: 
user activity varies in time 
and the load in one BS 
might have an effect 
on the BSs 
in its vicinity 
or those apart 
but otherwise connected  
(such as through 
a fast transportation 
connection) BSs.
When making 
short-term predictions, 
the impact of correlations 
is confined to the adjacent BSs.
In this study 
we account for this phenomenon 
by the incoming vehicular flow and speed.

Mathematically, we denote 
by $\mathbf{x}^\tau \in \mathcal{R}^p$
a random variable 
comprising historical measurements 
of $p$ metrics relevant to a given cell: 
cellular traffic volume 
or cellular load and 
road traffic volume 
and speed measurements. 
Given a time sequence of 
$M$ such historical observations 
$\{ \mathbf{x}^{\tau - M + 1}, \dots, \mathbf{x}^\tau \}$,  
our objective is to learn a prediction model $\mathcal{F}$ 
that can forecast the future call load
$\hat{x}$ in the cell during the next time step:
$\hat{x}^{\tau + 1} = \mathcal{F} (\mathbf{x}^{\tau - M + 1}, \dots, \mathbf{x}^\tau), \label{eq:focus}$
so that the prediction error $	L(\hat{x}, x)$ is minimized. 
The loss function $L(\mathord{\cdot})$ measures the difference between 
the estimated $\hat{x}$ and observed $x$ mobile cellular traffic load. 


\section{What about Data?} 
\label{sec:what_about_data}

The latest advancements 
in enhancing existing 
and designing new 
learning methods for 
mobile cellular traffic forecasting 
are surveyed 
in \cite{jiang2022cellular}
and \cite{dl2025ntp}.
In  Section~\ref{sec:cell_data}, 
we explore 
the state-of-the-art
from the perspective 
of data
because 
the existing surveys
do not question what data
would suit best
the purposes of 
the prediction task.
 In Section~\ref{sec:CDR},
 we turn our attention to 
how the data 
inherently generated 
and saved by 
mobile operators
for billing, 
management and 
optimization tasks---primarily, 
the call detail records 
(CDRs)---have been 
used so far.
We build upon 
the same premise as 
these studies---not only 
the human circadian 
cycle but virtually 
all human activities
are imprinted in 
the mobile cellular
use,~Section~\ref{sec:approach}.

\subsection{Cellular Forecasting: Data}
\label{sec:cell_data}

\subsubsection{Cellular KPIs}

Mobile cellular network operators
are reticent about sharing 
the data collected in their networks
because of privacy 
concerns\footnote{The 
	relevance
	of open 
	data sets for
	making 
	technological and 
	scientific progress,
	together with the 
	publicly available
	mobile cellular 
	network data
	are discussed in 
	\cite{amini2023cellular}.}.
This limitation
has shaped 
to a large extent 
the type of data used 
in the published research.
The Telecom Italia 
Big Data Challenge  
\cite{TelecomItalia2015}
is open access and
thereby, 
the predominantly 
used data set
(even nowadays
\cite{fu2025multi, ma2025mobimixer, ma2026sim}). 
It comprises 
calls, SMS 
and Internet traffic.
Therefore, 
the majority of
the deep learning
contributions continue
to measure cellular load
with the number of
calls.

Studies that fuel
their models with 
private mobile cellular
data are
yet rare.
Among them, the majority 
uses the volume of data 
to measure the BS load 
on a BS.
The fluctuating radio environment,
 can be modeled by
incorporating radio KPIs
into the data, \cite{okic2021pi}. 
Alternatively, the load in a cell 
is measured by  physical resource blocks
\cite{bega2019deepcog}.

\begin{table}[htbp]
	\caption{Use of Call Detail Records.}
	\begin{center}
		\begin{tabular}{l|c}
			\hline
			\textbf{Focus on / Purpose} &  \textbf{Studies}  \\
			\hline
			Dynamic population distribution modeling & \cite{wang2015understanding,chinamobile2016bigdata,calabrese2010real}, \\
			& \cite{gonzalez2008understanding,deville2014dynamic,tan2025spatiotemporal} \\
			& \cite{xu2018uncovering, HelsinkiUni22} \\
			Understanding daily flow of people & \cite{becker2011tale, becker2013human}  \\
			Computing urban mobile landscapes & \cite{pulselli2008computing} \\
			Monitoring real-time urban population density  & \cite{deville2014dynamic} \\
			Defining a city & \cite{dong2024defining} \\
			Effect of major events on human mobiltiy & \cite{zanella2022impact, bouzaghrane2024human, osorio2024analyzing} \\
			Functional role of urban parks & \cite{zanella2025digital} \\
			Urban inequalities & \cite{xu2025using} \\
			Road traffic estimation & \cite{becker2013human, herrera2010evaluation, becker2011route, review2008trafficestimates} \\
			Impact of commuting on rush hour & \cite{jarv2012mobile} \\
			\hline
		\end{tabular}
	\end{center}
	\label{tab:cdr}
\end{table}

\subsubsection{PoI}

In a series of articles
\cite{wang2015understanding, 
	xu2016understanding, 
	chinamobile2016bigdata,
	gong2024kgda},
the traffic patterns 
of thousands of cellular towers
in a large urban environment 
(Shanghai)
are distilled into 4 
basic temporal patterns.
It is shown that each 
traffic profile
can be mapped 
to a specific type
of urban region 
and hence,  
the region
can reveal 
the traffic profile 
of the serving BS.
PoI (such as museums, 
 hospitals or other 
 urban centers)
express the  context
of their geographical 
location
and their distribution 
is representative for 
the area 
whithin which
a BS operates.
This together with the conclusion
that  the BS's surroundings
are crucial exogenous factor
that shape the BS's traffic profile
 \cite{gong2024kgda}  
has motivated the inclusion of
PoI information
in the forecasting process:
\cite{feng2018deeptp}
and \cite{chai2025uomo}
are an earlier and 
a contemporary 
instances of 
the use of PoI 
for mobile data prediction.

\subsubsection{KGs} 

Recently introduced into
 cellular traffic 
forecasting by
\cite{KGgong2023empowering}
 and \cite{gong2024kgda},
KGs~\cite{hogan2021knowledge}
empower learning models 
with a comprehensive view on
the urban contex.
A spatial KG~\cite{KGgong2023empowering}
models the static environment 
of the BSs. 
Specifically, it embeds 
BSs spatial information,
proximity, 
functional,
(cellular traffic) pattern, 
and 
(cellular) volume and fluctuation rate
similarity between BSs.
This spatial KG is augmented 
to urban KG 
 \cite{KGgong2023empowering, gong2024kgda}
by 
combining expert knowledge
and urban specifics:
information on regions
(urban spaces delineated by
major road transportation paths),
business areas (clusters 
of economic and social activities),
PoI and categories 
as well as with information on
 spatial, subordinate and 
functional relationships 
between entities.
In order to learn from
its rich information,
a knowledge graph 
representation learning 
module is designed 
\cite{KGgong2023empowering, gong2024kgda}
to generate representations 
of the BSs that capture 
the network's spatial structure
 and nodes' functionality.
Similar to these two works, 
\cite{KG2025sttf} incorporates
semantic relationships between
BSs.

In summary, PoI and KGs
aim to enrich 
the cellular time series input
with information on 
the static BS environment.

\subsection{CDRs: Use}
\label{sec:CDR}

The mobile phones have been regarded as 
the best agent to monitor 
human mobility traces 
and the cellular networks---as sensing platforms 
of unprecedented scale, 
high resolution and low cost.
The assumption 
that there is a strong correlation
between population dynamics 
and mobile cellular traffic
is initially validated by 
ground truth data in 
\cite{chinamobile2016bigdata, calabrese2010real,  xu2018uncovering,   
	becker2013human, herrera2010evaluation, review2008trafficestimates,
	jarv2012mobile, tiru2007mobile, ranjan2012call,blondel2015survey}.
Contemporary studies  
\cite{tan2025spatiotemporal, dong2024defining}
point out
the high mobile phone use 
and nearly 100~\% 
global cellular coverage
as the basis for inferring 
metrics of interest
from CDRs.
Furthermore, a study  
of population fluctuations
in Helsinki metropolitan area
\cite{HelsinkiUni22} 
shows that 
census surveys 
under- or over-estimate 
population density 
during times of the day 
and locations in the city
when contrasted with 
estimates from CDRs.
Since the mobile cellular networks
reliably capture  
the dynamic human behaviour,  
 CDRs are seen as 
universal, consistent, and standardized data 
that serves as
a proxy to the total population activities---density 
and dynamics in an area 
as well as  socio-economic interaction. 
This explains their varied use, Table~\ref{tab:cdr},
in the development of sustainable and resilient cities.
Among these studies, 
\cite{ becker2013human, herrera2010evaluation, becker2011route, review2008trafficestimates} 
show that the mobile cellular systems
can be an alternative to 
transportation monitoring systems,
as road traffic parameters
can be estimated from cellular data,
which also reveals
the interrelation between 
vehicular and cellular traffic.

\begin{figure*}[htbp]
	\centerline{\includegraphics[width=0.98\textwidth]{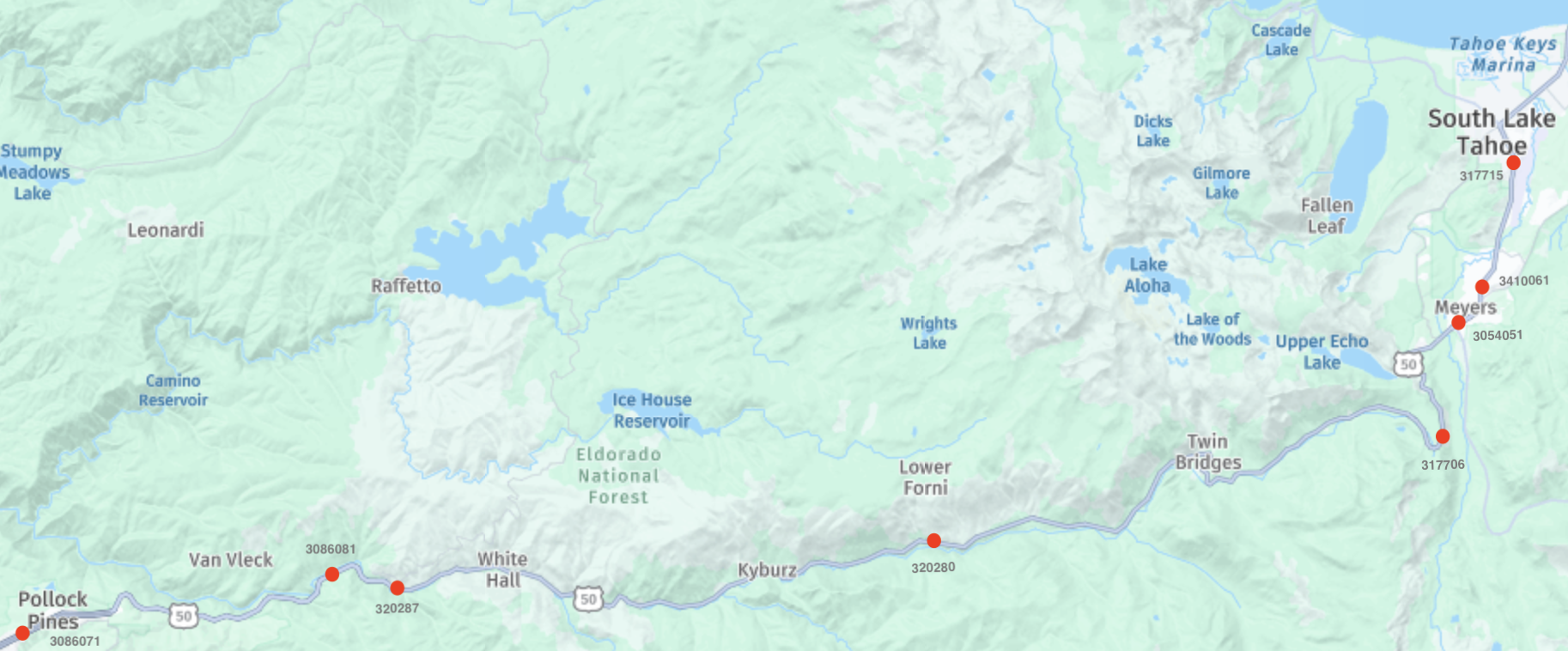}}
	\caption{A map of the section of US50-E El Dorado County freeway used in the study (in gray). The red dots indicate the approximate location of the PeMS detectors on that segment, from Pollock Pines to Lake Tahoe Airport. }
	\label{fig:map}
\end{figure*}

\begin{table*}[htbp]
	\caption{PeMS US50-E detectors used in the study, their specifications   
		and the distance between consecutive ones. \\
		Base stations (BSs) are denoted with the same identification number as the PeMS detectors.}
	\begin{center}
		\begin{tabular}{l|c|c|c|c|r}
			\hline
			\textbf{PeMS detectors} &  terrain  & population & design speed  & road  width & distance / BS \\
			&				&					&	limit in mph   &   in ft			&  range in miles \\
			\hline
			Mainline VDS \textbf{3086071}--50EB JWO Sly Park EB & mountainous & rural & 70  & 24  & 7.67 \\
			Mainline VDS \textbf{3086081}--50EB at Riverton Barn CCTV & mountainous & rural & 60  & 24   & 1.62   \\
			Mainline VDS \textbf{320287}--50EB at Ice House & mountainous & rural & 50  & 12   & 13.27   \\
			Mainline VDS \textbf{320280}--Wrights Lake Rd & mountainous & rural & 50  & 11   & 13.41   \\
			Mainline VDS \textbf{317706}--Echo Summit & mountainous & rural & 50  & 12   & 3.89   \\
			Mainline VDS \textbf{3054051}--50EB into Luther 50/89 R. & mountainous & rural & 40  & 12  & 0.99 \\
			Mainline VDS  \textbf{3410061}--50EB JEO Pioneer Trl & rolling & rural &  40  & 12   & 3.22 \\
			Mainline VDS  \textbf{317715}--F St & flat & urban & 40  & 32   & 1.46 \\
			\hline
		\end{tabular}
	\end{center}
	\label{tab:pems}
\end{table*}

\subsection{Employing Population and Cellular Traffic Time-Series}
\label{sec:approach}

\subsubsection{Our perspective}

We base our forecasting approach 
on the premise that there is 
a strong correlation between 
the presence of 
mobile cellular users
and the communication load placed on
the mobile cellular network. 
This view is explored extensively 
in  population dynamics research 
(summarized in Section~\ref{sec:CDR}).
Data are
a vital component of any
 data learning process.
Therefore, our prime interest is 
 in employing input  
 from which the learning model
 can gain insights into 
 the core processes that 
 generate  the
 mobile cellular traffic.
 Specifically, in addition to cellular KPIs,
 we incorporate  estimates of 
 the fluctuating number of 
 mobile cellular load sources.

\textbf{Contrasting with existing prior art.}
Among the published
deep learning 
contributions
devoted to 
mobile data 
forecasting,
there are few 
 we share 
the concept of 
modeling population 
dynamics with. 
\cite{handover20} and
\cite{fang2022sdgnet} 
explicitly try to model 
users' movement
based on the supposition 
that handover rates 
can capture to some extent
user mobility 
and traffic dynamics.
The combined effect of 
the devised model 
 \cite{fang2022sdgnet}
 and the handover frequencies
 is evaluated  
 \cite{handover20, fang2022sdgnet},
but the pure effect
of incorporating handover 
rates
is not quantified. 
In our work
\cite{vesselinova2025data} 
we do examine
the impact of 
handover rates
inclusion in the input 
on prediction accuracy. 
However, our results
are not conclusive
as they do not show 
consistent performance improvement. 
In \cite{chai2025uomo},
user movement 
is not predicted 
but traced via
the number of active 
mobile phone users
(a KPI available at BSs).
The ablation study 
in \cite{chai2025uomo}
shows that 
the number of 
active users 
better expresses the
dynamic characteristics 
of the mobile traffic and 
that this variable is more critical for
mobile data forecasting 
than the static 
PoI distribution
information.
This result corraborates our hypothesis.

\subsubsection{Data}
\label{sec:data}

We examine a scenario of 
connected vehicles on a highway. 
The load on the BSs 
that serve the highway 
is exclusively generated
by the road traffic on that highway 
(services used by the
autonomous driving vehicles 
and their passengers). 
To model the highway's 
population dynamics,
we employ two road variables---vehicular 
flow and speed---on each segment of the highway.
These are measured on 
a real highway, 
see Section~\ref{sec:highway} 
and Figure~\ref{fig:map}.
Due to the limitations 
of collecting  cellular data,
we simulate cellular KPIs
following the methodology 
developed in \cite{vesselinova2025data},
see Section~\ref{sec:input}.
Below we discuss the profile of the 
simulated cellular load time-series
and the correlation between 
the variables used to simulate 
the mobile cellular load 
and the simulated load.

\begin{figure*}[htbp!]
	\includegraphics[width=.25\textwidth]{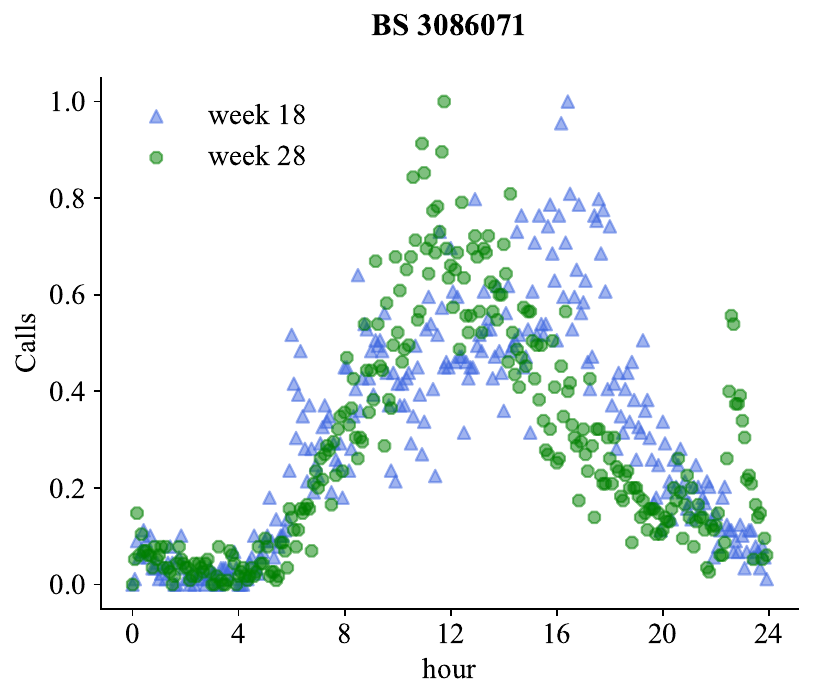}\includegraphics[width=.25\textwidth]{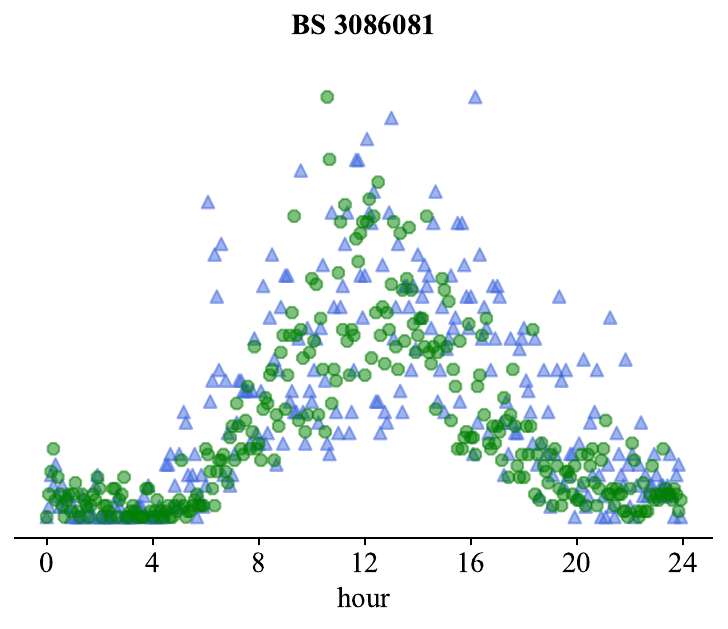}\includegraphics[width=.25\textwidth]{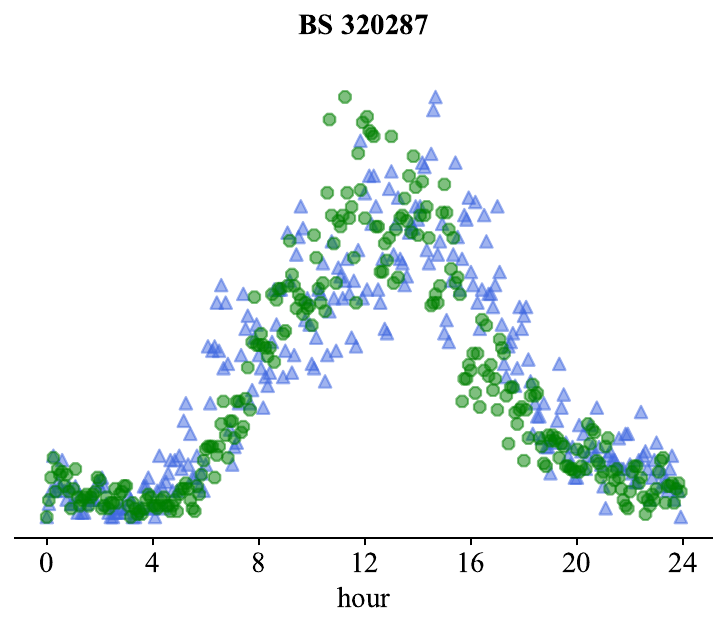}\includegraphics[width=.25\textwidth]{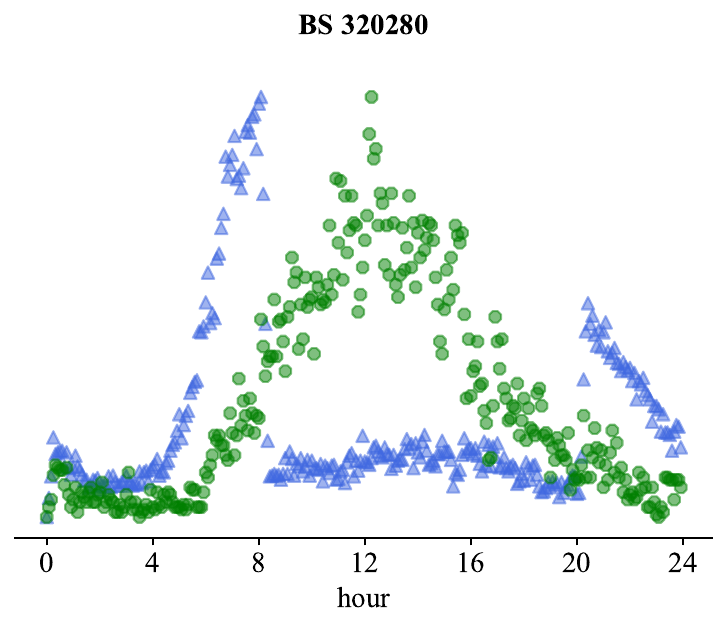}\newline\newline\includegraphics[width=.25\textwidth]{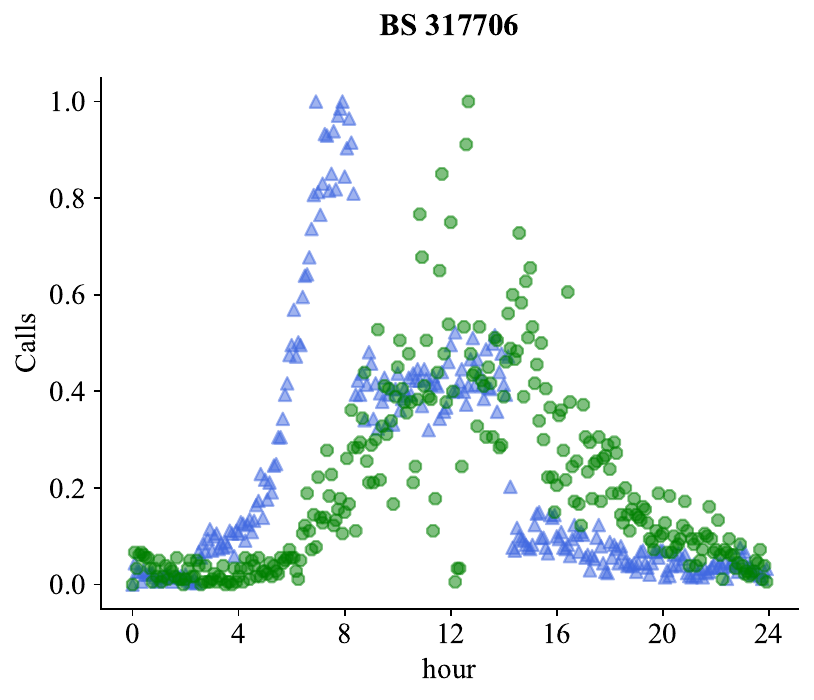}\includegraphics[width=.25\textwidth]{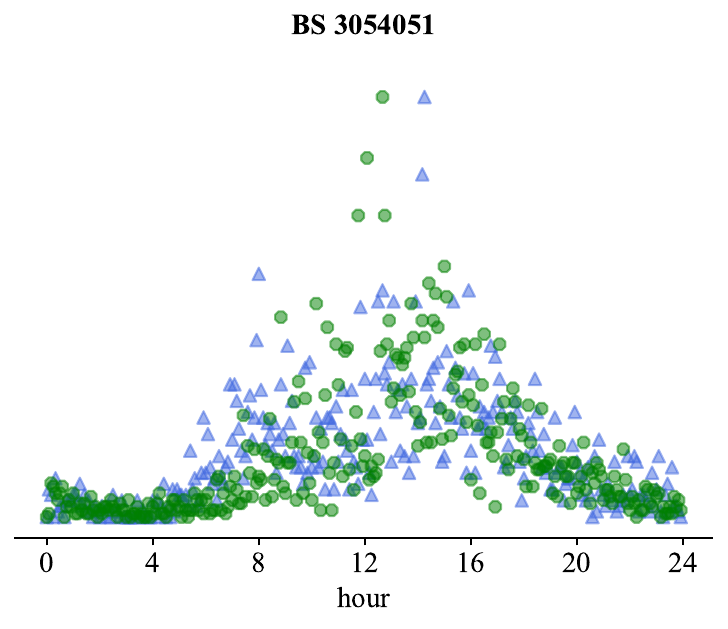}\includegraphics[width=.25\textwidth]{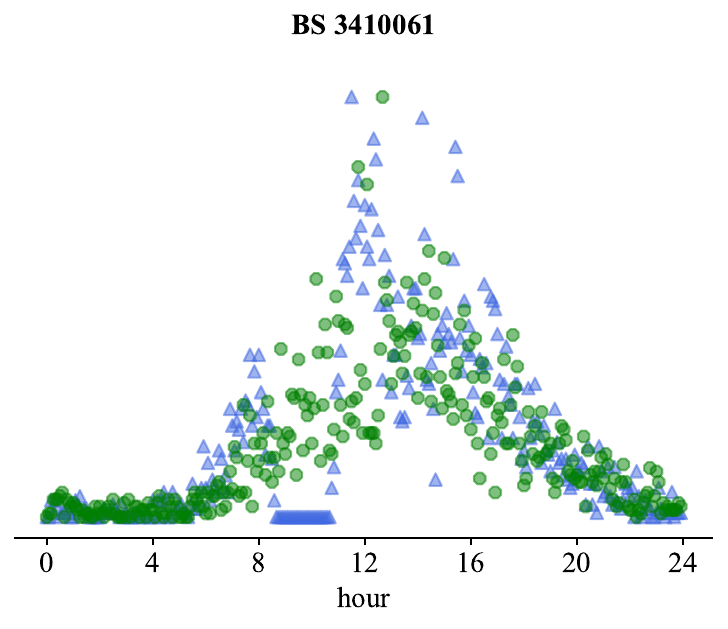}\includegraphics[width=.25\textwidth]{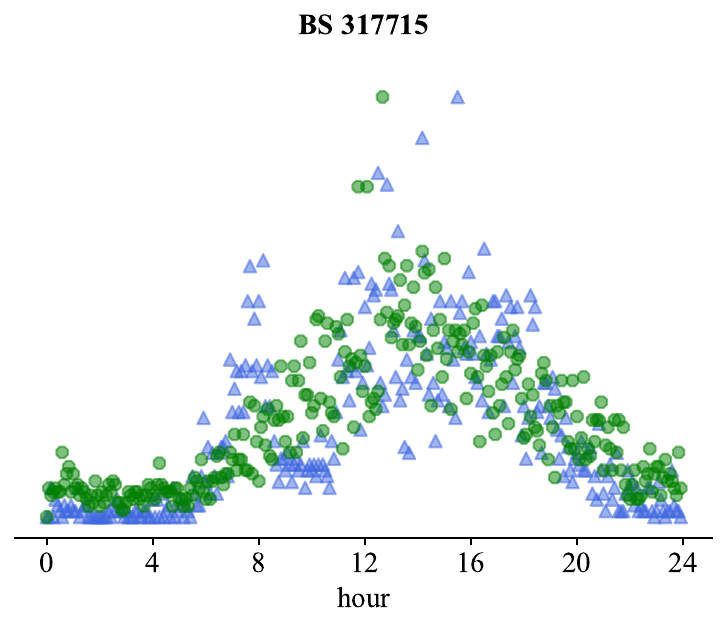}
	\caption{Normalized cellular load on Mondays during weeks 18 and 28.}
	\label{fig:normalized-calls-weeks}
\end{figure*}  

\begin{figure*}[htbp]
	\centerline{\includegraphics[width=.50\textwidth]{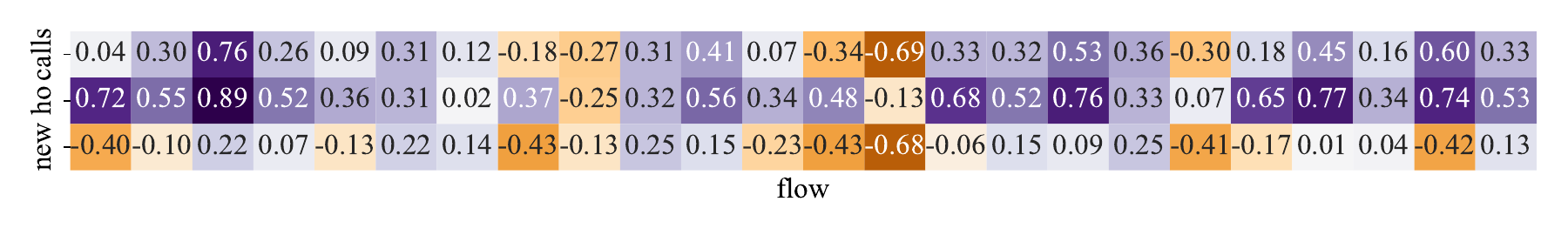} \includegraphics[width=.5\textwidth]{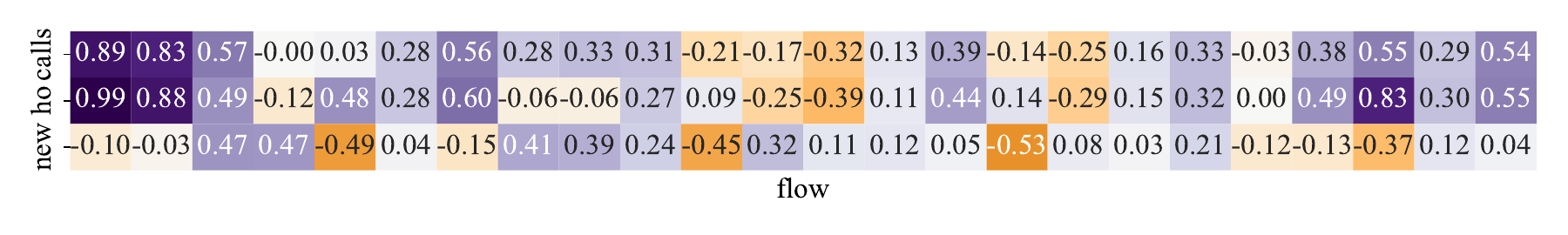}}
	\caption {Correlation between flow and new calls, handover calls, and (total number of) calls on Tuesdays, between 4 pm and 5 pm for weeks 13 to 36 at (\textbf{left})~BS~320280 and (\textbf{right})~BS~3054051.}
	\label{fig:correlation}
\end{figure*}

\textbf{Profile.}
The road data for the studied segment 
of the highway and time period,
shows highly fluctuating 
traffic volumes over space and time.
We emulate the vehicles' arrivals 
by a random process (Poisson)
during each 5-minute time slot~\cite{vesselinova2025data}
as the measured vehicular flow is aggregated
over 5-minute time windows.
For each vehicle,
the call arrivals 
are simulated by
a random probabilistic process
(Poisson~\cite{vesselinova2025data}) too.
The resulting BS's call volumes 
fluctuate sharply across short
and long time scales.
This is in-line with empirical data
from urban scenarios where 
the greatest peak-valley mobile data ratio 
is observed in the city's transport hubs 
when contrasted with other 
urban functional regions~\cite{wang2015understanding, xu2016understanding}. 
Figure~\ref{fig:normalized-calls-weeks}
depicts the normalized cellular load 
on Mondays during two different weeks
for each of the BSs providing coverage 
to the highway's segment.
Cellular traffic profiles and volumes 
often differ vastly between BSs
(location) as well as at the same BS,
day and time window but during 
different weeks. 
The large traffic variability in 
fine-grained (minutes and hours) and 
coarse-grained (days and weeks) time-scales,
and across BSs is evidenced in 
real mobile cellular networks in
cities~\cite{chinamobile2016bigdata}
and on highways~\cite{okic2021pi}.

\textbf{Correlation.}
The correlation between 
the different variables 
fluctuates substantially 
in time and 
between the different segments 
of the highway: 
strongly inverse, reciprocal 
or lacking any linear association.
Figure~\ref{fig:correlation} unveils that 
the vehicular--cellular
linear association is 
non-deterministic
along the 24-week period
even at the same location,
day and time.
At BS~320287, the correlation
between flow and calls 
during the 4~pm -- 5~pm
hour is 
$\rho=0.76$ in week~15,
$\rho=-0.69$  in week~26, 
and $\rho\approx0$  in weeks
17 and 24, for instance.
Likewise, at BS 3054051
it could vary from 
$\rho\approx0.9$ 
through $\rho=0$ 
to $\rho=-0.32$.

\begin{figure*}[htbp!]
	\includegraphics[width=.5\textwidth]{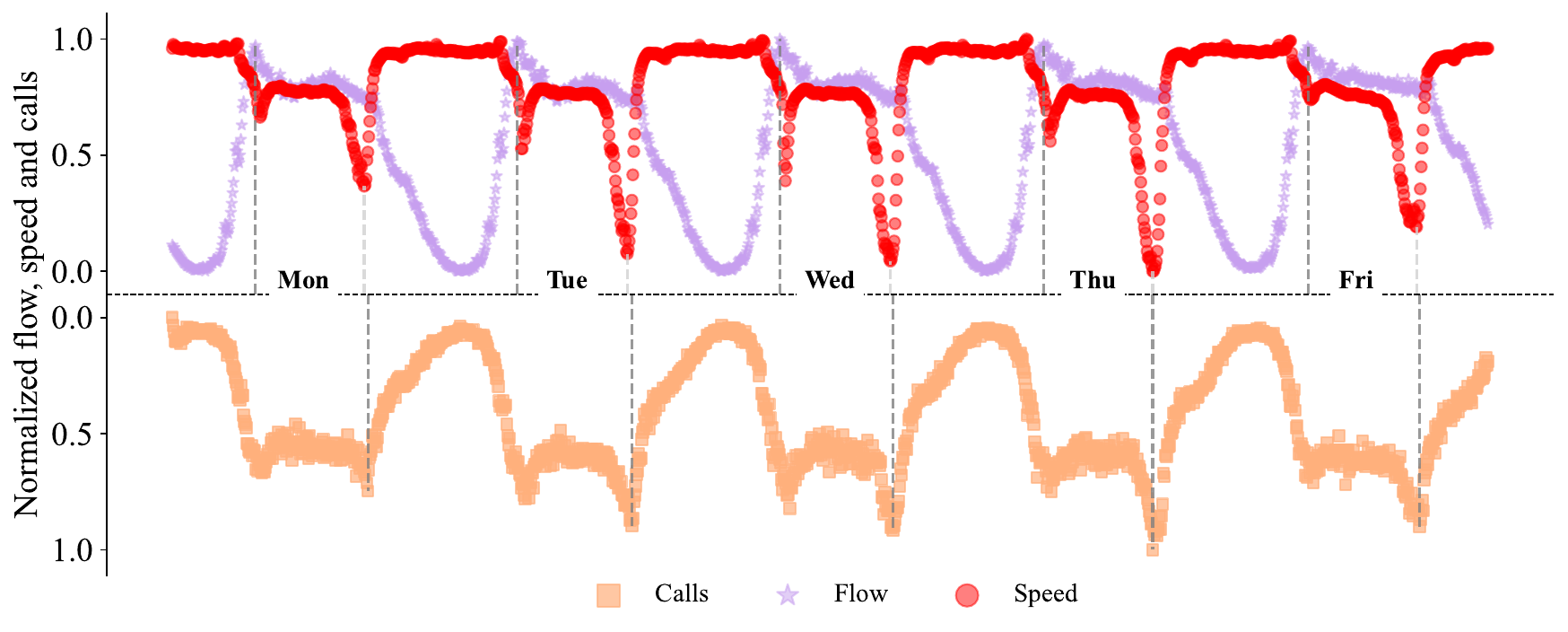}\includegraphics[width=.5\textwidth]{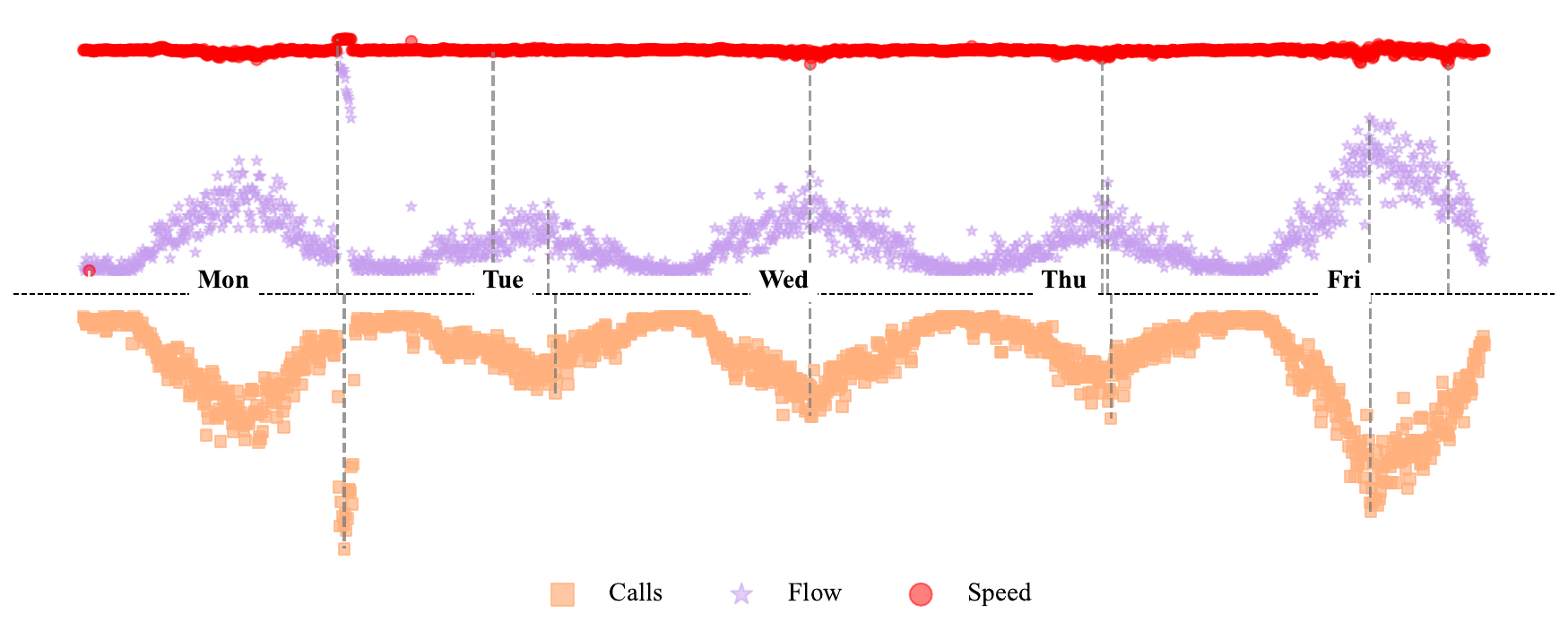}\newline\newline\includegraphics[width=.53\textwidth]{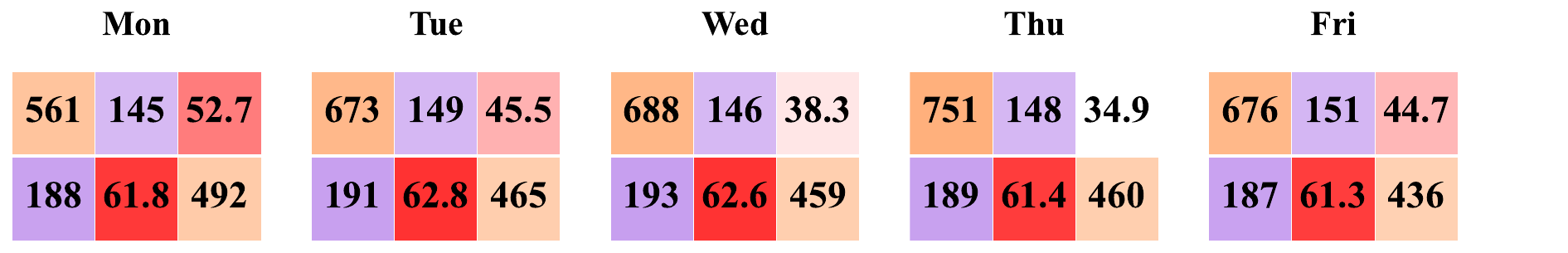}\includegraphics[width=.47\textwidth]{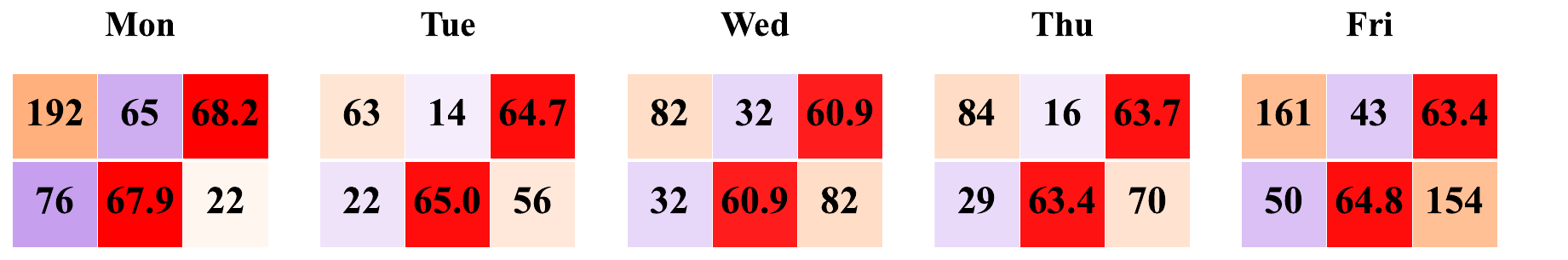}
	\caption {BS 320287 (\textbf{left}) week 15 and (\textbf{right}) week 16. (\textbf{top}) Flow, speed and calls. Calls are mirrored over the horizontal axis for at-a-glance silhouette recognition. (\textbf{bottom}) The maximum call load during a work day, followed by the value of the flow and speed at that instant of time (first row). The maximum flow during the same day and the corresponding average speed and calls volume (second row). The maximum daily call load is observed when the speed reaches its daily minimum during week 15. In contrast, during week 16---when the flow volumes are low compared to week 15---the speed does not have any perceptible effect on the flow nor calls. Then, the maximum call volume is defined by the maximum flow. }
	\label{fig:traffic_jam}
\end{figure*}

The flow--speed correlation 
depends on
the capacity of the road segment, 
vehicular flow, terrain, location and time.
Increased cellular volumes 
during road traffic incidents 
or jams are evidenced 
in practice~\cite{okic2021pi}.
Our data set truthfully reflects 
this  phenomenon too.
Figure~\ref{fig:traffic_jam} illustrates this.
The simulated call load
is inversely related to the speed
whenever the road gets congested, 
as the vehicle's dwell time on the road 
is prolonged and with it,
the probability of initiating a new call.

\textbf{Probabilistic processes.}
The generation of call requests
follows a probabilistic process.
Therefore, the cellular load 
would be different
even when the flow, speed,
and road conditions
remain the same. 
We highlight that 
the input to the learning model 
comprises the incoming new flow, 
not the total number of vehicles 
that would be 
in a cell
during a 5-minute time slot.
In contrast, the new load
on a BS (consisting of 
handed over and new calls) 
is determined not only by
the calls placed by 
the incoming vehicles 
but also by those 
that have arrived at the BS
during earlier time slots 
and are still served 
by that BS. 
This applies especially to 
BSs with long range. 
Conversely, when a BS
serves a short road segment---implying 
short dwell times
in the cell---not all
vehicles request service 
from that BS.

In summary, 
the sharply oscillating
load volumes and 
vastly different correlations
in time and space,
result from the~probabilistic processes 
that guide the vehicular flow and
the generation of cellular traffic.
These  make 
the prediction task 
quite challenging too.

\section{Experiments}
\label{sec:experiments}

\subsection{Setting}
\label{sec:setting}

\subsubsection{A highway scenario}
\label{sec:highway}

We consider a sector of the US50-E freeway in El Dorado County, California;
specifically, the one shown in Figure~\ref{fig:map}.
The California Department of Transportation  
(Caltrans) detectors, whose measurements are used in the study,
are listed in Table~\ref{tab:pems} in sequential order
together with the distance between 
each two consecutive detectors. 

The density of the PeMS detectors
seems to be guided by
similar principles---long-term 
density of the vehicular traffic---as 
the placement of BSs---population density 
and the particularities of the terrain.
Thereby, we choose 
the range of the BSs 
to match the distance 
between sensors. 
This provides us 
with a variety of 
use cases. 
The road segments 
exhibit diverse capacity, 
average vehicular density and 
propensity for congestion, 
which conditions together with 
the different BSs' ranges
model dynamic, time- and location-varying call volumes. 
For simplicity, we make the assumption that 
the vehicular traffic is unidirectional,
flowing from Pollac Pines to South Lake Tahoe.

\subsubsection{Input variables' values}
\label{sec:input}

Real road traffic data 
from Caltrans loop detectors 
on the US50-E highway 
are used in the experiments. 
The considered time period is 24-week-long,
from March 28, 2022 to September 9, 2022 
(week 13 to week 36 of 2022).
The data comprise 
the weekly flow and speed 
from Mondays to Fridays.
There are 288 data points 
per day due to 
the 5-minute granularity.

We generate the network statistics following the  methodology
developed in~\cite{vesselinova2025data} and input values as follows.
For each vehicle, new calls are generated with arrival rate $\lambda=1/5,$ 
one call per 5-minute interval on average, 
according to a Poisson process.
When defining the mean call duration and its variance, 
we are guided by~\cite{willkomm2008primary} and \cite{guo2007estimate}.
We set the mean call duration 
to $\mu=1$ minute for exponentially distributed calls and 
to $\mu_1=1$ and $\mu_2=10$~minutes 
for a mixture of two log-normally distributed calls, 
each with equal weight. 
The variance of the log-normal calls is 3 times larger 
than their mean~\cite{willkomm2008primary},
namely $\sigma_{1}^2 = 3$ and $\sigma_2^2=30$, correspondingly.
The speed is discretized as in~\cite{vesselinova2023road}.
The range of the  BSs is listed in Table~\ref{tab:pems}.
The  data are visualized and 
explored in the preceding Section~\ref{sec:data}.

\subsection{Machine learning model}
\label{sec:learning}

\subsubsection{Choice}The classic 
long short-term memory (LSTM) model 
has been broadly applied to time-series forecasting 
either as a singular learning structure or 
as a basic component of more complex models.
Furthermore, LSTM is shown to excel 
in very short-term 
cellular forecasting~\cite{fang2018mobile}. 

\subsubsection{Structure}The model 
we construct is composed by 
an LSTM layer, followed by 
a fully connected feedforward neural network 
(a multilayer perceptron).
This dense layer consists of a single unit  
and no activation function. 
It is a linear transformation that maps 
the final LSTM state
into a single target value---the 
estimated number of calls~in~a~BS.
Mathematically, the model 
can be represented 
as two functions $f(g),$ where
$g(\cdot)$ is the LSTM learning structure,
which transforms the input  data into new features.
The representation function $f(\cdot)$ 
maps the learned features into 
a cellular traffic load prediction 
for the BS under consideration.

\subsubsection{Implementation}Our model has a single LSTM layer with 16 cells.
In the training phase, 
we use the root mean square propagation 
and mean squared error (MSE) loss 
for optimizing the models parameters.
We set the length of the historical series 
to six samples (30~min). 
The prediction horizon is the next 5-minute interval. 
The 24 weeks of data are split 12:6:6 chronologically 
into training, validation, and testing
after which data are shuffled.

\subsection{Evaluation}
\label{sec:protocol}

\subsubsection{Methodology}
We assess 
the efficacy of 
our proposed approach 
by training the learning model 
with historical cellular data
\textit{together with} versus \textit{without} 
incorporating road traffic data.
Since our goal is to assess the effectiveness
of employing data intrinsic to
mobile cellular volume generation,
we contrast the prediction performance
of the implemented machine learning model 
when cellular KPIs are used (the baseline) 
with the model's performance 
when these same data
are enriched with 
vehicular flow and speed metrics.
We evaluate the predictions over
a comprehensive set of
 highway and cellular conditions.
Similar methodology 
for assessing the impact of 
data on prediction accuracy
is followed in~\cite{KGgong2023empowering}
when examining the efficacy of 
incorporating exogenous data
and in \cite{chai2025uomo}
when examining the impact of 
PoI information and 
the effect of the number of 
active mobile users 
on the forecasting performance.  

\subsubsection{Protocol}
We measure the 
forecasting performance with 
the typically used 
mean absolute error (MAE), 
mean absolute percentage error (MAPE), 
MSE and root mean squared error (RMSE).
To have a common basis for comparison and 
to fairly attribute 
any performance deviations
to the data set employed in learning,
we train and evaluate the model 
with the same~hyper-parameters~Section~\ref{sec:learning}
using data from the same period and location
but with different features: 
either containing purely network metrics
or a set of network and road traffic~metrics.
We measure the improvement in prediction 
as the percentage difference in error by
($ErrM_{net} - ErrM_{net\&road}) / ErrM_{net}$,
where $ErrM$ is the selected error measure.

\section{Results}
\label{sec:results}

We perform evaluations 
to answer 
the research questions:
\begin{itemize}
	\item\textbf{RQ1}: What is the impact of  data on performance? 
	\item\textbf{RQ2}: Do vehicular flow and average speed reliably model 
	population dynamics on highways? 
	\item\textbf{RQ3}: How flow estimation errors affect predictions? 
\end{itemize}

\setlength{\tabcolsep}{2.3pt}	
\begin{table*}[htbp]
	\caption{Prediction Performance on Calls Dataset and Flow, Speed and Calls Dataset\\
		A Mixture of Two Log-Normally Distributed Call Duration Times  \\ 24 Weeks, 12:6:6 Data Split }
	\begin{center}
		\scriptsize{
			\begin{tabular}{|c|lcr|lcr|lcr|lcr|lcr|lcr|lcr|lcr|}
				\hline
				\textbf{ }&\multicolumn{12}{c|}{\textbf{Calls}} &\multicolumn{12}{c|}{\textbf{Flow Speed Calls}} \\
				\cline{2-25} 
				\textbf{BS} 
				& \multicolumn{3}{c|}{\textbf{MAE}} 
				& \multicolumn{3}{c|}{\textbf{MSE}} 
				& \multicolumn{3}{c|}{\textbf{MAPE}} 
				& \multicolumn{3}{c|}{\textbf{RMSE}} 
				& \multicolumn{3}{c|}{\textbf{MAE}} 
				& \multicolumn{3}{c|}{\textbf{MSE}} 
				& \multicolumn{3}{c|}{\textbf{MAPE}} 
				& \multicolumn{3}{c|}{\textbf{RMSE}} \\
				& min & mdn & max & min & mdn & max & min & mdn & max & min & mdn & max 
				& min & mdn & max & min & mdn & max & min & mdn & max & min & mdn & max \\
				\hline
				\textbf{3086071} & .282 & \textbf{.283} & .287 & .154 & \textbf{.155} & .157 & 820.8 & \textbf{837.2} & 880.6 & .392 & \textbf{.393} & .396 & .187 & \textbf{.187} & .187 & .067 & \textbf{.067} & .068 & 506.8 & \textbf{515.6} & 523.9 & .259 & \textbf{.259} & .261 \\
				\textbf{3086081} & .394 & \textbf{.396} & .399 & .317 & \textbf{.318} & .319 & 335.2 & \textbf{347.4} & 365.4 & .563 & \textbf{.564} & .565 & .339 & \textbf{.340} & .342 & .237 & \textbf{.238} & .238 & 293.2 & \textbf{298.3} & 305.9 & .487 & \textbf{.487} & .488 \\
				\textbf{320287}  & .066 & \textbf{.067} & .067 & .009 & \textbf{.009} & .009 & 86.70 & \textbf{87.0} & 88.30 & .094 & \textbf{.094} & .094 & .051 & \textbf{.052} & .054 & .005 & \textbf{.006} & .006 & 64.60 & \textbf{66.60} & 67.70 & .073 & \textbf{.075} & .076 \\
				\textbf{320280}  & .080 & \textbf{.081} & .097 & .013 & \textbf{.013} & .015 & 54.00 & \textbf{54.60} & 57.30 & .112 & \textbf{.114} & .121 & .065 & \textbf{.066} & .070 & .008 & \textbf{.008} & .009 & 47.30 & \textbf{47.7} & 49.10 & .090 & \textbf{.091} & .094 \\
				\textbf{317706}  & .155 & \textbf{.156} & .158 & .058 & \textbf{.059} & .060 & 127.5 & \textbf{128.4} & 137.8 & .241 & \textbf{.242} & .244 & .129 & \textbf{.13} & .135 & .038 & \textbf{.039} & .040 & 107.4 & \textbf{110.8} & 112.0 & .196 & \textbf{.198} & .199 \\
				\textbf{3054051} & .434 & \textbf{.434} & .439 & .485 & \textbf{.492} & .503 & 116.0 & \textbf{118.0} & 128.3 & .697 & \textbf{.702} & .709 & .364 & \textbf{.369} & .377 & .341 & \textbf{.347} & .354 & 84.3 & \textbf{90.10} & 93.10 & .584 & \textbf{.589} & .595 \\
				\textbf{3410061} & .404 & \textbf{.404} & .407 & .389 & \textbf{.391} & .397 & 696.4 & \textbf{725.7} & 757.9 & .624 & \textbf{.626} & .63 & .354 & \textbf{.356} & .358 & .309 & \textbf{.31} & .316 & 484.8 & \textbf{494.5} & 503.2 & .556 & \textbf{.557} & .562 \\
				\textbf{317715}  & .321 & \textbf{.324} & .325 & .242 & \textbf{.243} & .246 & 103.5 & \textbf{109.3} & 117.3 & .492 & \textbf{.493} & .496 & .295 & \textbf{.297} & .300 & .208 & \textbf{.210} & .218 & 87.40 & \textbf{92.40} & 92.80 & .456 & \textbf{.459} & .467 \\
				\hline
				\multicolumn{25}{l}{$^{\mathrm{a}}$The three values listed per error type are the minimum, median and maximum. The MAPE values are in percentage.}
			\end{tabular}
		}
		\label{tab:mixture}
	\end{center}
\end{table*}

\begin{table*}[htbp]
	\caption{Prediction Performance on Calls Dataset and Flow, Speed and Calls Dataset \\
		Exponentially Distributed Call Duration Time  \\ 24 Weeks, 12:6:6 Data Split } 
	\begin{center}
		\scriptsize{
			\begin{tabular}{|c|lcr|lcr|lcr|lcr|lcr|lcr|lcr|lcr|}
				\hline
				\textbf{ }&\multicolumn{12}{c|}{\textbf{Calls}} &\multicolumn{12}{c|}{\textbf{Flow Speed Calls}} \\
				\cline{2-25} 
				\textbf{BS} 
				& \multicolumn{3}{c|}{\textbf{MAE}} 
				& \multicolumn{3}{c|}{\textbf{MSE}} 
				& \multicolumn{3}{c|}{\textbf{MAPE}} 
				& \multicolumn{3}{c|}{\textbf{RMSE}} 
				& \multicolumn{3}{c|}{\textbf{MAE}} 
				& \multicolumn{3}{c|}{\textbf{MSE}} 
				& \multicolumn{3}{c|}{\textbf{MAPE}} 
				& \multicolumn{3}{c|}{\textbf{RMSE}} \\
				& min & mdn & max & min & mdn & max & min & mdn & max & min & mdn & max
				& min & mdn & max & min & mdn & max & min & mdn & max & min & mdn & max \\
				\hline
				\textbf{3086071} & .282 & \textbf{.283} & .283 & .151 & \textbf{.152} & .152 & 2085 & \textbf{2108} & 2113 & .389 & \textbf{.389} & .389 & .187 & \textbf{.187} & .189 & .066 & \textbf{.066} & .066 & 1347 & \textbf{1377} & 1388 & .257 & \textbf{.257} & .257 \\
				\textbf{3086081} & .396 & \textbf{.398} & .402 & .315 & \textbf{.315} & .318 & 103.9 & \textbf{105.0} & 106.4 & .561 & \textbf{.561} & .564 & .327 & \textbf{.331} & .334 & .217 & \textbf{.218} & .222 & 87.5 & \textbf{88.0} & 93.7 & .466 & \textbf{.467} & .471 \\
				\textbf{320287}  & .058 & \textbf{.058} & .059 & .007 & \textbf{.007} & .007 & 55.5 & \textbf{56.3} & 58.3 & .082 & \textbf{.082} & .084 & .041 & \textbf{.041} & .042 & .003 & \textbf{.003} & .004 & 42.4 & \textbf{43.0} & 43.5 & .058 & \textbf{.059} & .060 \\
				\textbf{320280}  & .073 & \textbf{.082} & .086 & .011 & \textbf{.012} & .012 & 38.3 & \textbf{38.6} & 39.9 & .103 & \textbf{.109} & .111 & .049 & \textbf{.053} & .055 & .005 & \textbf{.005} & .005 & 26.2 & \textbf{27.1} & 28.7 & .068 & \textbf{.070} & .071 \\
				\textbf{317706}  & .159 & \textbf{.162} & .164 & .064 & \textbf{.066} & .069 & 127.4 & \textbf{132.2} & 135.7 & .254 & \textbf{.258} & .262 & .100 & \textbf{.103} & .109 & .024 & \textbf{.027} & .038 & 80.30 & \textbf{82.10} & 84.70 & .154 & \textbf{.164} & .194 \\
				\textbf{3054051} & .483 & \textbf{.493} & .500 & .584 & \textbf{.600} & .607 & 129.7 & \textbf{135.9} & 141.6 & .764 & \textbf{.775} & .779 & .317 & \textbf{.319} & .336 & .231 & \textbf{.234} & .246 & 82.50 & \textbf{85.30} & 92.80 & .480 & \textbf{.484} & .496 \\
				\textbf{3410061} & .398 & \textbf{.402} & .408 & .335 & \textbf{.343} & .349 & 104.6 & \textbf{109.7} & 116.9 & .579 & \textbf{.586} & .59 & .295 & \textbf{.297} & .300 & .187 & \textbf{.188} & .189 & 72.5 & \textbf{73.0} & 73.6 & .432 & \textbf{.433} & .434 \\
				\textbf{317715}  & .270 & \textbf{.274} & .278 & .147 & \textbf{.151} & .152 & 108.0 & \textbf{110.1} & 113.0 & .384 & \textbf{.388} & .389 & .207 & \textbf{.207} & .210 & .086 & \textbf{.087} & .087 & 77.10 & \textbf{77.60} & 80.40 & .294 & \textbf{.295} & .296 \\			
				\hline
			\end{tabular}
		}
		\label{tab:exp}
	\end{center}
\end{table*}

\begin{figure*}[htbp]
	\centerline{\includegraphics[width=.45\textwidth]{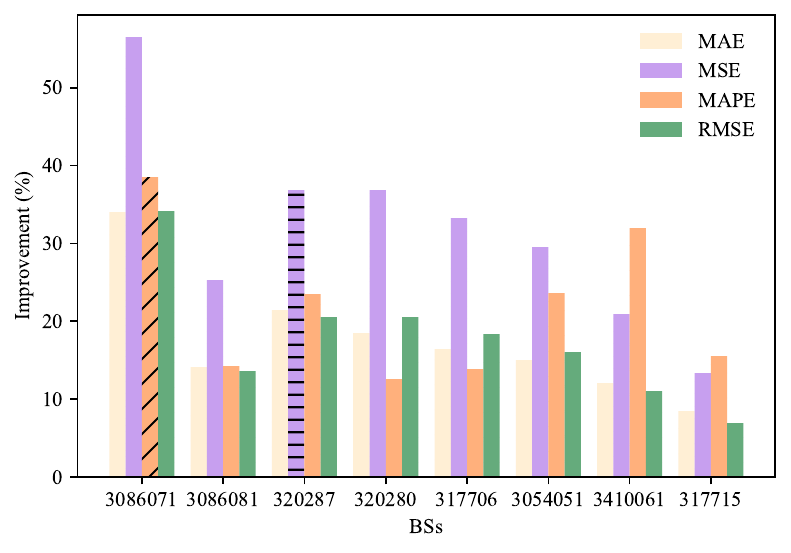} \includegraphics[width=0.45\textwidth]{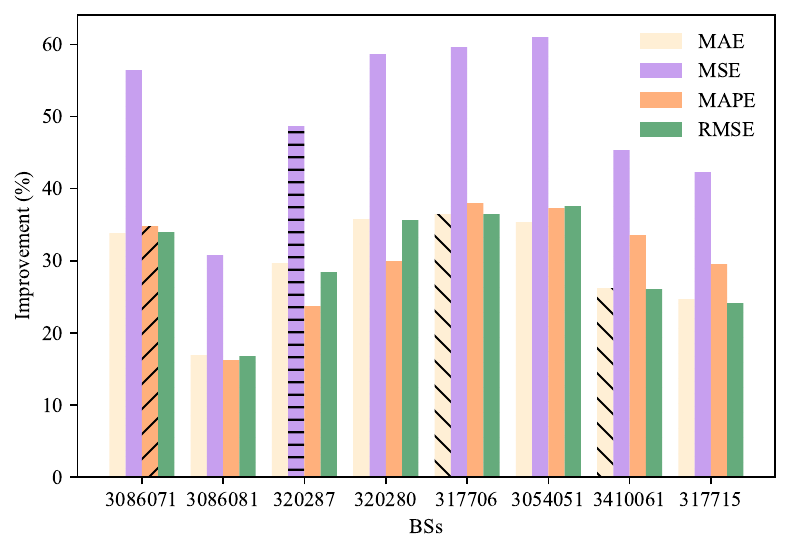}}
	\caption{Improvement in prediction when employing calls vs calls, flow and speed data: (\textbf{left})~a mixture of two log-normally distributed calls  and (\textbf{right})~exponentially distributed call duration.}
	\label{fig:improvement}
\end{figure*}

\subsection{Performance}
\label{sec:performance}

\subsubsection{Main outcome}

Comprehensive results from  
four different data sets
are summarized in 
Table~\ref{tab:mixture}---Table~\ref{tab:noise}.
The call duration is modeled by 
1)~a mixture of two log-normally distributed random variables
or by 2)~an exponentially distributed random variable. 
For each distribution we compile two data sets:
1)~calls (denoting total number of calls---the sum of new and handover calls) only; 
and 2)~flow, speed and calls.

To convey the shape and hence, provide a clearer picture on 
the distribution of errors, for each error metric 
Table~\ref{tab:mixture}---Table~\ref{tab:noise} 
show the minimum, median and maximum
of the 5 simulation runs conducted for each data set and BS.

\textbf{Employing data that captures the processes
		intrinsic to mobile cellular traffic generation
		consistently improves cellular load prediction performance (RQ1). }
	All BSs experience improvements in all their forecasting measures 
	when learning from both network and road traffic data.	
	The prediction error reduction among all BSs,
	when considering the case of log-normally distributed calls,
	is between $8.4\%$ and $33.9\%$~(MAE),
	$13.3\%$ and $56.5\%$~(MSE), $12.5\%$ and $38.4\%$~(MAPE), and 
	$6.9\%$ and $34\%$~(RMSE) when in addition to (total number of) calls, 
	also flow and speed are employed, and when comparing the medians of all BSs, 
	see Figure~\ref{fig:improvement} (left). 
	Similar improvement trends are recorded 
	for the minimum and maximum errors,
	Table~\ref{tab:mixture}. 
	For the exponentially distributed calls, 
	the decrease in error due to 
	use of population dynamics is 
	between $16.89\%$ and $35.74\%$~(MAE),
	$30.67\%$ and $61.02\%$~(MSE), 
	$16.16\%$ and $37.92\%$~(MAPE), and 
	$16.73\%$ and $37.57\%$~(RMSE)
	for the medians, see 
	Figure~\ref{fig:improvement} (right). 
	The trends remain similar 
	for the minimum and maximum
	errors.

\begin{table*}[htbp]
	\caption{Prediction Performance with Gaussian Noise Added to the Flow \\
		A Mixture of Two Log-Normally Distributed Call Duration Times (left) \\
		and Exponentially Distributed Call Duration Time (right) \\ 24 Weeks, 12:6:6 Data Split} 
	\begin{center}
		\scriptsize{
			\begin{tabular}{|c|lcr|lcr|lcr|lcr|lcr|lcr|lcr|lcr|}
				\hline
				\textbf{ }&\multicolumn{12}{c|}{\textbf{Flow Speed Calls}} &\multicolumn{12}{c|}{\textbf{Flow Speed Calls}} \\
				\cline{2-25} 
				\textbf{BS} 
				& \multicolumn{3}{c|}{\textbf{MAE}} 
				& \multicolumn{3}{c|}{\textbf{MSE}} 
				& \multicolumn{3}{c|}{\textbf{MAPE}} 
				& \multicolumn{3}{c|}{\textbf{RMSE}} 
				& \multicolumn{3}{c|}{\textbf{MAE}} 
				& \multicolumn{3}{c|}{\textbf{MSE}} 
				& \multicolumn{3}{c|}{\textbf{MAPE}} 
				& \multicolumn{3}{c|}{\textbf{RMSE}} \\
				& min & mdn & max & min & mdn & max & min & mdn & max & min & mdn & max
				& min & mdn & max & min & mdn & max & min & mdn & max & min & mdn & max \\
				\hline
				\textbf{3086071} & .205 & \textbf{.205} & .206 & .077 & \textbf{.077} & .078 & 594.5 & \textbf{599.4} & 602.1 & .278 & \textbf{.278} & .279 & .204 & \textbf{.204} & .205 & .075 & \textbf{.075} & .076 & 1474 & \textbf{1483} & 1498 & .274 & \textbf{.274} & .275 \\
				\textbf{3086081} & .346 & \textbf{.349} & .35 & .244 & \textbf{.245} & .247 & 293.1 & \textbf{315.5} & 320.6 & .494 & \textbf{.495} & .497 & .336 & \textbf{.338} & .339 & .224 & \textbf{.225} & .226 & 87.9 & \textbf{88.60} & 91.00 & .474 & \textbf{.475} & .476 \\
				\textbf{320287}  & .057 & \textbf{.057} & .058 & .006 & \textbf{.006} & .006 & 71.70 & \textbf{72.90} & 74.80 & .079 & \textbf{.080} & .080 & .046 & \textbf{.047} & .048 & .004 & \textbf{.004} & .005 & 44.20 & \textbf{45.10} & 46.50 & .064 & \textbf{.065} & .068 \\
				\textbf{320280}  & .070 & \textbf{.072} & .074 & .009 & \textbf{.009} & .010 & 50.5 & \textbf{50.90} & 51.10 & .096 & \textbf{.097} & .098 & .060 & \textbf{.061} & .066 & .006 & \textbf{.007} & .007 & 30.90 & \textbf{31.50} & 32.20 & .079 & \textbf{.081} & .084 \\
				\textbf{317706}  & .134 & \textbf{.136} & .138 & .040 & \textbf{.042} & .043 & 110.0 & \textbf{111.3} & 119.5 & .200 & \textbf{.205} & .207 & .108 & \textbf{.109} & .110 & .027 & \textbf{.027} & .028 & 86.3 & \textbf{87.60} & 88.7 & .163 & \textbf{.165} & .168 \\
				\textbf{3054051} & .369 & \textbf{.371} & .375 & .348 & \textbf{.354} & .358 & 84.0 & \textbf{85.9} & 87.8 & .590 & \textbf{.595} & .598 & .323 & \textbf{.324} & .333 & .235 & \textbf{.237} & .244 & 85.6 & \textbf{86.40} & 93.2 & .485 & \textbf{.487} & .493 \\
				\textbf{3410061} & .360 & \textbf{.361} & .365 & .313 & \textbf{.318} & .327 & 481.2 & \textbf{495.0} & 514.7 & .56 & \textbf{.564} & .572 & .306 & \textbf{.307} & .316 & .195 & \textbf{.195} & .197 & 73.2 & \textbf{74.10} & 77.3 & .441 & \textbf{.442} & .444 \\
				\textbf{317715}  & .301 & \textbf{.303} & .306 & .214 & \textbf{.214} & .224 & 88.5 & \textbf{91.0} & 96.4 & .462 & \textbf{.463} & .473 & .216 & \textbf{.216} & .217 & .093 & \textbf{.093} & .094 & 78.7 & \textbf{80.00} & 81.0 & .304 & \textbf{.305} & .306 \\
				\hline
			\end{tabular}
		}
		\label{tab:noise}
	\end{center}
\end{table*}

\begin{figure*}[htbp]
	\centerline{\includegraphics[width=.45\textwidth]{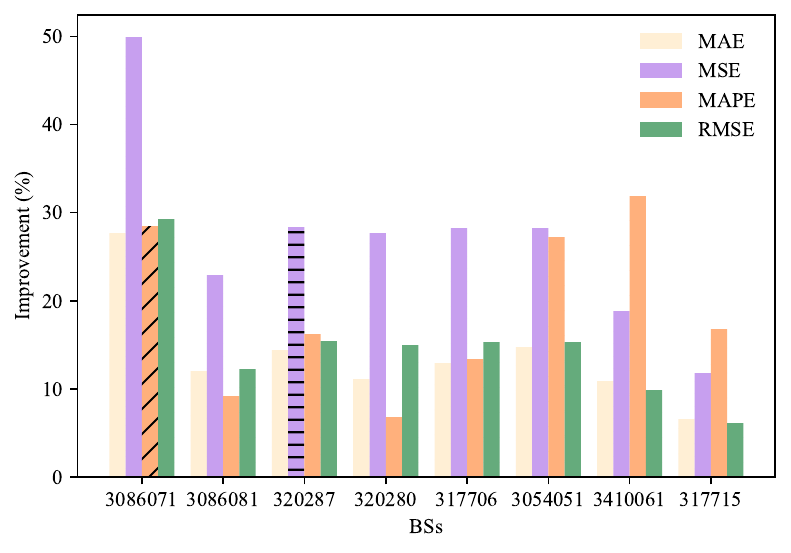} \includegraphics[width=0.45\textwidth]{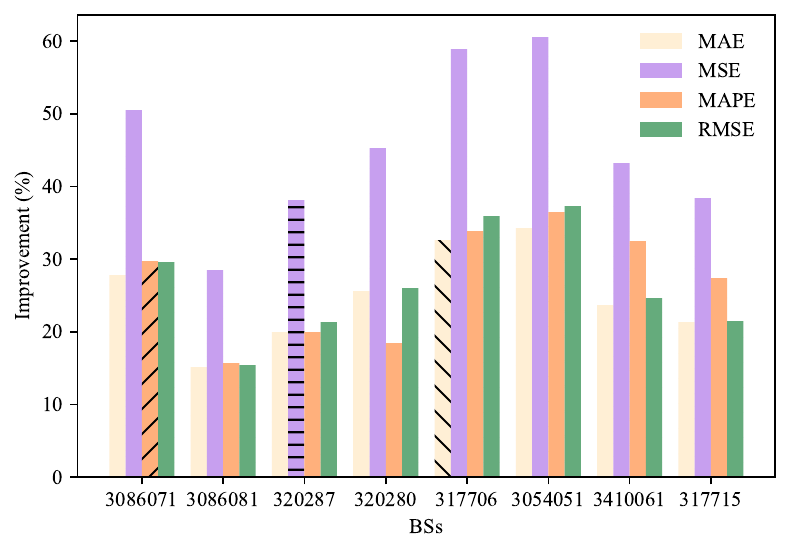}}
	\caption{Flow estimated with errors. Improvement in prediction when employing calls vs calls, flow estimated with errors  and speed data: (\textbf{left})~a mixture of two log-normally distributed calls  and (\textbf{right})~exponentially distributed call duration.}
	\label{fig:noise}
\end{figure*}

\textbf{Road traffic indicators can capture
		the processes underlying  
		mobile cellular load generation (RQ2).}
	We employ flow and speed as a means to characterize 
	the underlying population dynamics in highways 
	and through them the data generation process. 
	The flow captures the number of potential call generation sources. 
	The speed can indicate dynamics in the vehicular density
	and can also serve as a gauge for road traffic congestion; 
	hence, for increased cellular load.

\subsubsection{Sensitivity to road metrics accuracy (\textbf{RQ3})}

To assess the sensitivity of 
the mobile cellular predictions 
to estimation errors in the road predictions,
we introduce estimation errors 
in the flow measured by 
the PeMS detectors. 
Specifically, we assume 
a prediction error of $5\%$ 
in the number of vehicles 
per 5-minute intervals.
We model the error  
by adding 
Gaussian noise to 
the real PeMS measurements
$	\hat{v} =  v + \epsilon,$  $\epsilon \sim \mathcal{N}(0,\sigma),$ 
where $v$ denotes 
the real PeMS flow data 
and $\hat{v}$  is the estimated flow.
Whereas the input to our learning model
accounts for the estimated value 
of the flow variable $\hat{v}$,
we generated 
the mobile cellular traffic load 
with the true PeMS 
measurement $v$ value. 

The results are reported in Table~\ref{tab:noise} and show that 
although the noise in the flow variable decreases cell load prediction accuracy,
employing road data remains largely beneficial. 
Overall, across all error measures 
and all BSs and the 2 data sets, the prediction improvement 
when considering the mixture of the log-normally distributed calls 
is between $9.20\%$ and $49.94\%$,  
Figure~\ref{fig:noise}~(left), 
and for the exponentially distributed call duration, it is 
between $15.07\%$ and $60.57\%$,  
Figure~\ref{fig:noise}~(right),
for the medians when using estimated (namely, with errors) 
road traffic time-series data.

\begin{figure*}[htbp]
	\centerline{\includegraphics[width=.24\textwidth]{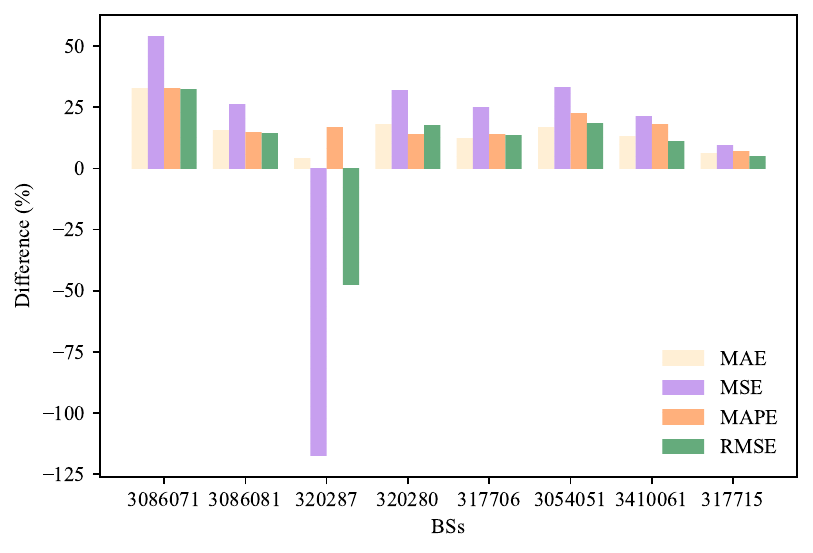} \includegraphics[width=0.24\textwidth]{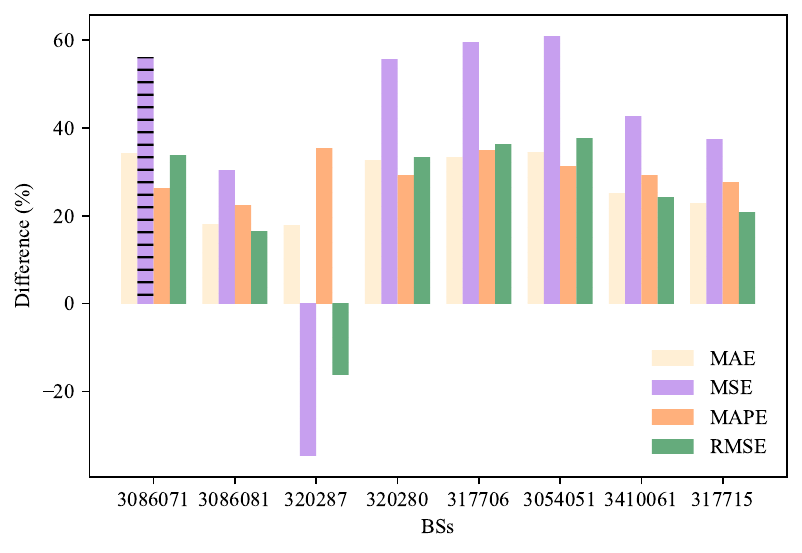}
		\includegraphics[width=.24\textwidth]{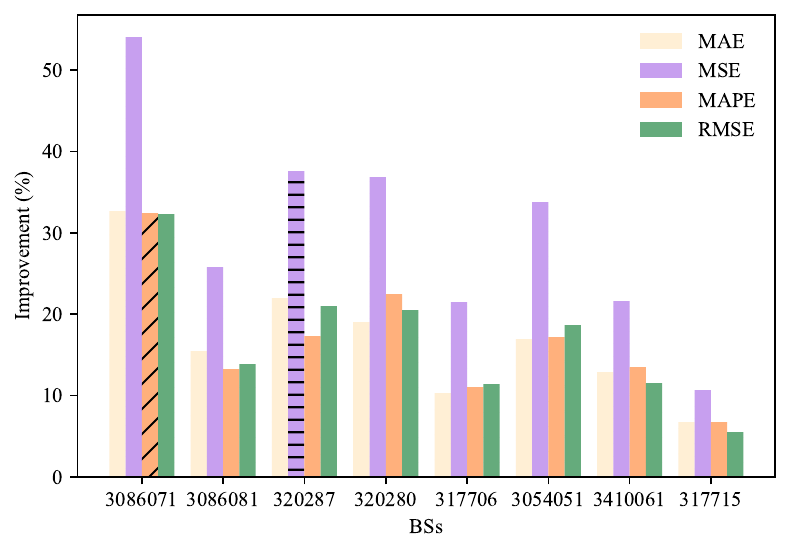} \includegraphics[width=0.24\textwidth]{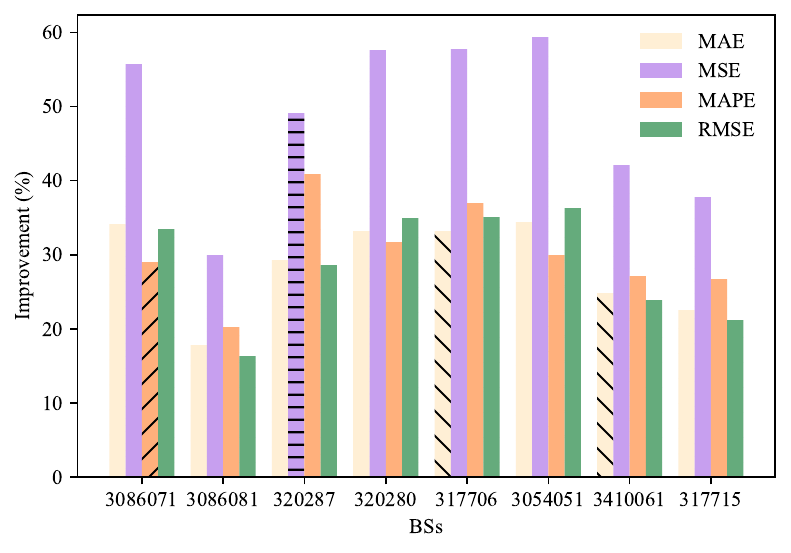}}
	\caption{Data---weeks: 27 to 33, split: 3:2:2 (first two) and 4:1:2 (last two plots). Difference in prediction performance when contrasting the homogeneous (calls) with heterogeneous (calls, flow and speed) data sets for a mixture of two log-normally distributed call duration times (first and third plots) and exponential call duration (second and last plots) per BS.}
	\label{fig:improvement_data_split}
\end{figure*}

\subsection{Analysis}

\subsubsection{Variability in BS performance}
 \label{sec:variability}

The BSs in the studied scenario 
differ in range (miles covered), 
exhibited call load,
observed handover statistics, 
capacity of the road segment 
served by a BS, 
experienced vehicular flow 
and measured average speed
as well as in the variables' 
daily and weekly patterns. 
Thereby, they also perform differently 
in terms of prediction accuracy. 
Notably, for all error measures 
but MAPE, the BSs can be 
ranked in the same order independent of 
whether a purely network data  
or a heterogeneous set combining 
road and call data 
is used for training\footnote{
When BSs are ranked based on their MAPE score, 
the order is different from the ranking 
when using the other three measures of error, 
yet it remains independent from data.}. 
This fact evinces 
a general trend---best to worst
performance is determined by 
an amalgam of cell characteristics 
and the phenomena that occur there. 
We improve with data
the model's prediction performance
on top of this trend.

We also look at the magnitude of improvement
brought by employing population dynamics,
Figure~\ref{fig:improvement}. 
For the exponential
call duration case,
the largest decrease in 
prediction error 
is observed for BS~3054051,
which exhibits the worst 
prediction performance
among the BSs.
It is also the BS with 
the shortest radio coverage.
When the call duration 
is modeled with 
a mixture of 
two log-normally distributed 
random variables,
the largest improvement
is experienced by 
BS~3086071,
which is among 
the least performing BSs
and the one without handovers
(as it is the first in our scenario).

The results from this study 
validate an observation 
made in~\cite{vesselinova2023road}---cells
with larger highway coverage
tend to achieve higher prediction accuracy. 
This can be intuitively explained by 
the vehicles' dwell time, which is longer
there and hence, the vehicles have 
a higher probability of placing
at least one call on their serving BS. 
Then, the number of vehicles
provides a lower bound on
the number of calls.
This might gauge 
the learning model
towards 
more accurate forecasting.

\subsubsection{Call distribution}

Employing flow and speed, leads to 
significant improvements in 
prediction capacity 
independent of
the call duration distribution.
However, the scale of improvement 
depends on the distribution.
The reduction in error is more substantial 
in the memoryless, 
exponential case (from $16.16\%$ to $61.02\%$) than 
in the mixture of two log-normally distributed variables
(from $6.89\%$ to $56.51\%$). 

\subsubsection{Data split}

The results summarized in Section~\ref{sec:performance} 
show the performance of the learning model
when trained with the first 12 weeks, 
validated on the subsequent 6 weeks
and tested on the last 6 weeks of the data set.
When the time series span
is shorter, in some instances
we observe
departures from the general trends
described in Section~\ref{sec:performance}.
We examine the performance of BS~320287 
when the data set consists of weeks~27--33 and 
the data is chronologically split~into 3:2:2 
for training, validation and testing.
Figure~\ref{fig:improvement_data_split}
illustrates the difference in 
the overall prediction performance 
for the two call distribution cases (first two plots).
Across the 10 test days, the overall 
MSE and RMSE prediction errors are larger 
when employing the heterogeneous data. 
However, when the same 
weeks 27--33 are partitioned 
into 4:1:2, 
the results are consistent with 
what we report above---decreased prediction errors
across all performance measures and BSs 
when employing flow and speed time-series
in addition to call time-series data,
Figure~\ref{fig:improvement_data_split}
(last two plots).
These results 
suggest that the learning model 
is not able to capture all of
the underlying patterns 
when retaining  
3 instead of 4 training weeks 
from the heterogeneous 
6-week data set.
A detailed look into the data 
and performance results
supports our understanding
as explained below. 

\setlength{\tabcolsep}{2.3pt}	
\begin{table*}[htbp]
	\caption{Prediction Performance on Calls Dataset and Flow, Speed and Calls Dataset\\
		A Mixture of Two Log-Normally Distributed Call Duration Times  \\  Weeks 27--33, 3:2:2 Data Split }
	\begin{center}
		\scriptsize{
			\begin{tabular}{|c|lcr|lcr|lcr|lcr|lcr|lcr|lcr|lcr|}
				\hline
				\textbf{ } & \multicolumn{12}{c|}{\textbf{Calls}} & \multicolumn{12}{c|}{\textbf{Flow Speed Calls}} \\
				\cline{2-25} 
				\textbf{week} 
				& \multicolumn{3}{c|}{\textbf{MAE}} 
				& \multicolumn{3}{c|}{\textbf{MSE}} 
				& \multicolumn{3}{c|}{\textbf{MAPE}} 
				& \multicolumn{3}{c|}{\textbf{RMSE}} 
				& \multicolumn{3}{c|}{\textbf{MAE}} 
				& \multicolumn{3}{c|}{\textbf{MSE}} 
				& \multicolumn{3}{c|}{\textbf{MAPE}} 
				& \multicolumn{3}{c|}{\textbf{RMSE}} \\
				& min & mdn & max & min & mdn & max & min & mdn & max & min & mdn & max
				& min & mdn & max & min & mdn & max & min & mdn & max & min & mdn & max \\
				\hline
				& .175 & \textbf{.177} & .179 & .057 & \textbf{.057} & .058 & 80.26 & \textbf{83.48} & 86.01 & .238 & \textbf{.239} & .241 & .132 & \textbf{.133} & .139 & .034 & \textbf{.035} & .036 & 66.55 & \textbf{68.63} & 73.06 & .186 & \textbf{.187} & .191 \\
				& .148 & \textbf{.149} & .151 & .041 & \textbf{.042} & .043 & 104.7 & \textbf{108.6} & 112.1 & .203 & \textbf{.204} & .207 & .110 & \textbf{.112} & .120 & .024 & \textbf{.024} & .028 & 88.81 & \textbf{93.29} & 97.39 & .153 & \textbf{.156} & .167 \\
				\textbf{32} & .154 & \textbf{.155} & .158 & .046 & \textbf{.046} & .047 & 91.65 & \textbf{94.61} & 95.78 & .214 & \textbf{.215} & .218 & .122 & \textbf{.124} & .127 & .028 & \textbf{.028} & .030 & 66.88 & \textbf{68.8} & 78.20 & .167 & \textbf{.168} & .175 \\
				& .168 & \textbf{.17} & .173 & .052 & \textbf{.052} & .053 & 92.48 & \textbf{94.71} & 99.42 & .228 & \textbf{.229} & .231 & .133 & \textbf{.134} & .137 & .034 & \textbf{.035} & .036 & 76.53 & \textbf{79.30} & 81.27 & .184 & \textbf{.188} & .190 \\
				& .233 & \textbf{.235} & .236 & .100 & \textbf{.100} & .101 & 66.15 & \textbf{68.57} & 70.95 & .316 & \textbf{.317} & .318 & .\textcolor{red}{295} & \textcolor{red}{\textbf{.363}} & \textcolor{red}{.517} & \textcolor{red}{.304} & \textcolor{red}{\textbf{.522}} & \textcolor{red}{1.25} & 42.39 & \textbf{44.4} & 54.12 & \textcolor{red}{.551} & \textcolor{red}{\textbf{.723}} & \textcolor{red}{1.12} \\
				\hline
				& .148 & \textbf{.149} & .152 & .041 & \textbf{.041} & .042 & 67.54 & \textbf{68.85} & 69.86 & .202 & \textbf{.203} & .206 & .113 & \textbf{.114} & .121 & .024 & \textbf{.025} & .027 & 64.83 & \textbf{68.58} & 74.26 & .155 & \textbf{.157} & .166 \\
				& .157 & \textbf{.158} & .160 & .046 & \textbf{.047} & .048 & 81.47 & \textbf{84.03} & 86.09 & .215 & \textbf{.216} & .218 & .119 & \textbf{.122} & .127 & .026 & \textbf{.027} & .031 & 44.22 & \textbf{45.03} & 56.28 & .162 & \textbf{.164} & .176 \\
				\textbf{33} & .143 & \textbf{.144} & .147 & .039 & \textbf{.040} & .041 & 107.3 & \textbf{110.0} & 113.0 & .198 & \textbf{.199} & .201 & .120 & \textbf{.121} & .125 & .028 & \textbf{.029} & .030 & 102.6 & \textbf{104.3} & 108.3 & .168 & \textbf{.170} & .173 \\
				& .181 & \textbf{.182} & .184 & .062 & \textbf{.062} & .063 & 78.36 & \textbf{80.00} & 80.95 & .248 & \textbf{.249} & .251 & .146 & \textbf{.147} & .150 & .038 & \textbf{.039} & .041 & 73.63 & \textbf{76.04} & 79.25 & .195 & \textbf{.197} & .202 \\
				& .217 & \textbf{.219} & .220 & .087 & \textbf{.087} & .088 & 45.37 & \textbf{46.07} & 46.62 & .294 & \textbf{.295} & .297 & .190 & \textbf{.209} & .234 & .078 & \textcolor{red}{\textbf{.114}} & \textcolor{red}{.206} & 33.29 & \textbf{34.83} & 40.27 & .279 & \textcolor{red}{\textbf{.337}} & \textcolor{red}{.454} \\
				\hline				
				\multicolumn{25}{l}{$^{\mathrm{a}}$ Test results from Mondays to Fridays (in sequential order) for weeks 32 and 33.}
			\end{tabular}
		}
		\label{tab:mixture_split_322}
	\end{center}
\end{table*}

\setlength{\tabcolsep}{2.3pt}	
\begin{table*}[htbp]
	\caption{Prediction Performance on Calls Dataset and Flow, Speed and Calls Dataset\\
		A Mixture of Two Log-Normally Distributed Call Duration Times  \\  Weeks 27--33, 4:1:2 Data Split }
	\begin{center}
		\scriptsize{
			\begin{tabular}{|c|lcr|lcr|lcr|lcr|lcr|lcr|lcr|lcr|}
				\hline
				\textbf{ } & \multicolumn{12}{c|}{\textbf{Calls}} & \multicolumn{12}{c|}{\textbf{Flow Speed Calls}} \\
				\cline{2-25} 
				\textbf{week} 
				& \multicolumn{3}{c|}{\textbf{MAE}} 
				& \multicolumn{3}{c|}{\textbf{MSE}} 
				& \multicolumn{3}{c|}{\textbf{MAPE}} 
				& \multicolumn{3}{c|}{\textbf{RMSE}} 
				& \multicolumn{3}{c|}{\textbf{MAE}} 
				& \multicolumn{3}{c|}{\textbf{MSE}} 
				& \multicolumn{3}{c|}{\textbf{MAPE}} 
				& \multicolumn{3}{c|}{\textbf{RMSE}} \\
				& min & mdn & max & min & mdn & max & min & mdn & max & min & mdn & max
				& min & mdn & max & min & mdn & max & min & mdn & max & min & mdn & max \\
				\hline
				& .170 & \textbf{.174} & .177 & .055 & \textbf{.056} & .056 & 88.55 & \textbf{94.76} & 98.32 & .234 & \textbf{.236} & .237 & .129 & \textbf{.131} & .133 & .032 & \textbf{.033} & .033 & 77.82 & \textbf{82.80} & 85.40 & .180 & \textbf{.182} & .183 \\
				& .146 & \textbf{.147} & .148 & .040 & \textbf{.040} & .041 & 85.92 & \textbf{87.32} & 89.50 & .200 & \textbf{.201} & .202 & .109 & \textbf{.110} & .113 & .023 & \textbf{.024} & .024 & 72.42 & \textbf{75.18} & 76.78 & .152 & \textbf{.154} & .155 \\
				\textbf{32} & .152 & \textbf{.153} & .154 & .045 & \textbf{.045} & .045 & 91.99 & \textbf{94.76} & 97.52 & .211 & \textbf{.212} & .212 & .109 & \textbf{.110} & .113 & .023 & \textbf{.024} & .024 & 72.42 & \textbf{75.18} & 76.78 & .152 & \textbf{.154} & .155 \\
				& .165 & \textbf{.167} & .169 & .050 & \textbf{.051} & .052 & 93.59 & \textbf{95.6} & 98.45 & .225 & \textbf{.226} & .228 & .120 & \textbf{.121} & .122 & .026 & \textbf{.027} & .027 & 66.69 & \textbf{68.86} & 73.68 & .162 & \textbf{.164} & .165 \\
				& .229 & \textbf{.231} & .233 & .096 & \textbf{.097} & .101 & 73.77 & \textbf{76.70} & 77.60 & .311 & \textbf{.312} & .318 & .166 & \textbf{.179} & .185 & .055 & \textbf{.064} & .072 & 33.71 & \textbf{38.60} & 42.88 & .234 & \textbf{.254} & .269 \\
				\hline
				& .145 & \textbf{.147} & .149 & .040 & \textbf{.040} & .040 & 79.71 & \textbf{80.98} & 83.07 & .199 & \textbf{.200} & .201 & .110 & \textbf{.112} & .114 & .023 & \textbf{.024} & .024 & 81.10 & \textbf{82.96} & 87.64 & .153 & \textbf{.154} & .156 \\
				& .154 & \textbf{.156} & .158 & .044 & \textbf{.045} & .046 & 93.32 & \textbf{95.97} & 98.69 & .210 & \textbf{.212} & .214 & .118 & \textbf{.119} & .121 & .025 & \textbf{.026} & .026 & 45.60 & \textbf{46.94} & 50.37 & .159 & \textbf{.16} & .162 \\
				\textbf{33} & .14 & \textbf{.141} & .144 & .038 & \textbf{.038} & .039 & 117.6 & \textbf{119.5} & 122.94 & .194 & \textbf{.196} & .196 & .116 & \textbf{.117} & .120 & .026 & \textbf{.027} & .028 & 114.7 & \textbf{119.5} & 124.7 & .163 & \textbf{.165} & .168 \\
				& .178 & \textbf{.179} & .181 & .059 & \textbf{.061} & .062 & 86.98 & \textbf{89.39} & 92.41 & .244 & \textbf{.247} & .249 & .142 & \textbf{.144} & .147 & .037 & \textbf{.037} & .040 & 77.28 & \textbf{81.77} & 88.06 & .192 & \textbf{.194} & .200 \\
				& .213 & \textbf{.216} & .218 & .083 & \textbf{.085} & .088 & 43.74 & \textbf{44.82} & 45.26 & .289 & \textbf{.292} & .296 & .171 & \textbf{.172} & .177 & .054 & \textbf{.055} & .062 & 30.54 & \textbf{32.67} & 33.55 & .233 & \textbf{.235} & .249 \\
				\hline				
				\multicolumn{25}{l}{$^{\mathrm{a}}$ Test results from Mondays to Fridays (in sequential order) for weeks 32 and 33.}
			\end{tabular}
		}
		\label{tab:mixture_split_412}
	\end{center}
\end{table*}

\setlength{\tabcolsep}{2.3pt}	
\begin{table*}[htbp]
	\caption{Prediction Performance on Calls Dataset and Flow, Speed and Calls Dataset\\
		Exponentially Distributed Call Duration Time,  Weeks 27--33, 3:2:2 Data Split }
	\begin{center}
		\scriptsize{
			\begin{tabular}{|c|lcr|lcr|lcr|lcr|lcr|lcr|lcr|lcr|}
				\hline
				\textbf{ } & \multicolumn{12}{c|}{\textbf{Calls}} & \multicolumn{12}{c|}{\textbf{Flow Speed Calls}} \\
				\cline{2-25} 
				\textbf{week} 
				& \multicolumn{3}{c|}{\textbf{MAE}} 
				& \multicolumn{3}{c|}{\textbf{MSE}} 
				& \multicolumn{3}{c|}{\textbf{MAPE}} 
				& \multicolumn{3}{c|}{\textbf{RMSE}} 
				& \multicolumn{3}{c|}{\textbf{MAE}} 
				& \multicolumn{3}{c|}{\textbf{MSE}} 
				& \multicolumn{3}{c|}{\textbf{MAPE}} 
				& \multicolumn{3}{c|}{\textbf{RMSE}} \\
				& min & mdn & max & min & mdn & max & min & mdn & max & min & mdn & max
				& min & mdn & max & min & mdn & max & min & mdn & max & min & mdn & max \\
				\hline
				& .180 & \textbf{.181} & .182 & .058 & \textbf{.059} & .059 & 77.12 & \textbf{79.47} & 80.86 & .241 & \textbf{.242} & .243 & .125 & \textbf{.129} & .134 & .029 & \textbf{.031} & .033 & 41.65 & \textbf{43.75} & 48.63 & .172 & \textbf{.175} & .181 \\
				& .152 & \textbf{.154} & .157 & .042 & \textbf{.043} & .044 & 84.33 & \textbf{85.64} & 89.31 & .206 & \textbf{.207} & .209 & .096 & \textbf{.103} & .109 & .017 & \textbf{.019} & .020 & 49.14 & \textbf{51.45} & 53.53 & .131 & \textbf{.137} & .142 \\
				\textbf{32} & .150 & \textbf{.150} & .153 & .040 & \textbf{.041} & .041 & 108.0 & \textbf{110.3} & 112.8 & .201 & \textbf{.201} & .203 & .105 & \textbf{.108} & .113 & .020 & \textbf{.021} & .023 & 75.10 & \textbf{77.62} & 81.90 & .141 & \textbf{.143} & .152 \\
				& .184 & \textbf{.184} & .185 & .063 & \textbf{.063} & .064 & 78.22 & \textbf{79.23} & 82.32 & .251 & \textbf{.252} & .253 & .130 & \textbf{.131} & .137 & .031 & \textbf{.032} & .035 & 53.51 & \textbf{56.89} & 61.35 & .176 & \textbf{.179} & .187 \\
				& .231 & \textbf{.232} & .238 & .103 & \textbf{.105} & .111 & 47.75 & \textbf{49.02} & 51.25 & .322 & \textbf{.323} & .333 &  \textcolor{red}{.300 }&  \textcolor{red}{\textbf{.348}} & .484 & \textcolor{red}{.362} & \textcolor{red}{\textbf{.518}} & \textcolor{red}{1.28} & 35.50 & \textbf{36.49} & 42.50 & \textcolor{red}{.602} & \textcolor{red}{\textbf{.718}} & \textcolor{red}{1.13} \\
				\hline
				& .150 & \textbf{.151} & .153 & .042 & \textbf{.042} & .042 & 42.88 & \textbf{43.76} & 45.855 & .204 & \textbf{.205} & .206 & .101 & \textbf{.105} & .109 & .021 & \textbf{.022} & .023 & 29.77 & \textbf{30.64} & 31.48 & .145 & \textbf{.149} & .153 \\
				& .161 & \textbf{.162} & .164 & .050 & \textbf{.051} & .051 & 69.00 & \textbf{70.58} & 73.72 & .224 & \textbf{.225} & .226 & .113 & \textbf{.115} & .122 & .023 & \textbf{.024} & .028 & 48.41 & \textbf{51.55} & 60.67 & .152 & \textbf{.154} & .166 \\
				\textbf{33} & .154 & \textbf{.156} & .160 & .050 & \textbf{.050} & .051 & 126.2 & \textbf{127.9} & 131.7 & .223 & \textbf{.224} & .226 & .104 & \textbf{.106} & .12 & .021 & \textbf{.022} & .027 & 65.77 & \textbf{71.67} & 95.71 & .146 & \textbf{.148} & .165 \\
				& .188 & \textbf{.189} & .190 & .070 & \textbf{.070} & .071 & 126.2 & \textbf{128.6} & 131.1 & .264 & \textbf{.265} & .267 & .126 & \textbf{.130} & .142 & .031 & \textbf{.032} & .038 & 70.91 & \textbf{73.35} & 91.16 & .176 & \textbf{.180} & .194 \\
				& .228 & \textbf{.230} & .234 & .098 & \textbf{.099} & .103 & 46.92 & \textbf{48.63} & 51.74 & .313 & \textbf{.315} & .321 & .189 & \textbf{.200} & .244 & .098 & \textcolor{red}{\textbf{.123}} & \textcolor{red}{.211} & 36.46 & \textbf{38.96} & 42.24 & .313 & \textcolor{red}{\textbf{.351}} & \textcolor{red}{.460} \\
				\hline				
				\multicolumn{25}{l}{$^{\mathrm{a}}$ Test results from Mondays to Fridays (in sequential order) for weeks 32 and 33.}
			\end{tabular}
		}
		\label{tab:exp_split_322}
	\end{center}
\end{table*}

\setlength{\tabcolsep}{2.3pt}	
\begin{table*}[htbp]
	\caption{Prediction Performance on Calls Dataset and Flow, Speed and Calls Dataset\\
		Exponentially Distributed Call Duration Time,  Weeks 27--33, 4:1:2 Data Split }
	\begin{center}
		\scriptsize{
			\begin{tabular}{|c|lcr|lcr|lcr|lcr|lcr|lcr|lcr|lcr|}
				\hline
				\textbf{ } & \multicolumn{12}{c|}{\textbf{Calls}} & \multicolumn{12}{c|}{\textbf{Flow Speed Calls}} \\
				\cline{2-25} 
				\textbf{week} 
				& \multicolumn{3}{c|}{\textbf{MAE}} 
				& \multicolumn{3}{c|}{\textbf{MSE}} 
				& \multicolumn{3}{c|}{\textbf{MAPE}} 
				& \multicolumn{3}{c|}{\textbf{RMSE}} 
				& \multicolumn{3}{c|}{\textbf{MAE}} 
				& \multicolumn{3}{c|}{\textbf{MSE}} 
				& \multicolumn{3}{c|}{\textbf{MAPE}} 
				& \multicolumn{3}{c|}{\textbf{RMSE}} \\
				& min & mdn & max & min & mdn & max & min & mdn & max & min & mdn & max
				& min & mdn & max & min & mdn & max & min & mdn & max & min & mdn & max \\
				\hline
				& .177 & \textbf{.178} & .18 & .056 & \textbf{.057} & .058 & 125.79 & \textbf{128.5} & 134.7 & .236 & \textbf{.239} & .240 & .123 & \textbf{.125} & .127 & .029 & \textbf{.029} & .030 & 69.76 & \textbf{76.12} & 77.95 & .169 & \textbf{.171} & .173 \\
				& .151 & \textbf{.153} & .154 & .041 & \textbf{.042} & .042 & 74.333 & \textbf{75.73} & 77.65 & .203 & \textbf{.204} & .205 & .096 & \textbf{.099} & .102 & .017 & \textbf{.018} & .018 & 44.58 & \textbf{46.04} & 47.31 & .131 & \textbf{.133} & .136 \\
				\textbf{32}   & .146 & \textbf{.148} & .149 & .039 & \textbf{.039} & .040 & 172.2 & \textbf{180.5} & 189.0 & .197 & \textbf{.198} & .199 & .104 & \textbf{.106} & .109 & .019 & \textbf{.020} & .020 & 115.7 & \textbf{126.3} & 146.0 & .139 & \textbf{.141} & .143 \\
				& .180 & \textbf{.181} & .182 & .061 & \textbf{.061} & .062 & 88.81 & \textbf{91.97} & 96.33 & .246 & \textbf{.248} & .249 & .129 & \textbf{.130} & .131 & .030 & \textbf{.031} & .033 & 55.97 & \textbf{57.68} & 62.31 & .173 & \textbf{.176} & .181 \\
				& .229 & \textbf{.230} & .232 & .101 & \textbf{.102} & .103 & 44.81 & \textbf{45.82} & 46.85 & .318 & \textbf{.320} & .321 & .158 & \textbf{.165} & .171 & .051 & \textbf{.054} & .060 & 26.07 & \textbf{28.32} & 29.72 & .226 & \textbf{.233} & .245 \\
				\hline
				& .148 & \textbf{.149} & .151 & .040 & \textbf{.041} & .041 & 42.01 & \textbf{43.14} & 43.83 & .200 & \textbf{.202} & .203 & .100 & \textbf{.103} & .105 & .021 & \textbf{.021} & .022 & 29.06 & \textbf{30.21} & 31.873 & .145 & \textbf{.147} & .15 \\
				& .158 & \textbf{.159} & .161 & .048 & \textbf{.048} & .049 & 64.00 & \textbf{65.36} & 66.27 & .219 & \textbf{.220} & .222 & .112 & \textbf{.114} & .119 & .023 & \textbf{.023} & .024 & 47.45 & \textbf{50.60} & 53.34 & .150 & \textbf{.153} & .155 \\
				\textbf{33} & .152 & \textbf{.155} & .157 & .048 & \textbf{.049} & .049 & 345.5 & \textbf{351.7} & 357.0 & .220 & \textbf{.221} & .222 & .102 & \textbf{.103} & .104 & .021 & \textbf{.021} & .022 & 163.4 & \textbf{170.9} & 184.7 & .143 & \textbf{.144} & .147 \\
				& .185 & \textbf{.186} & .188 & .067 & \textbf{.068} & .068 & 253.9 & \textbf{257.5} & 260.7 & .258 & \textbf{.260} & .261 & .124 & \textbf{.126} & .130 & .030 & \textbf{.030} & .031 & 110.5 & \textbf{119.4} & 129.2 & .173 & \textbf{.174} & .177 \\
				& .226 & \textbf{.228} & .230 & .097 & \textbf{.097} & .099 & 72.21 & \textbf{77.98} & 81.22 & .311 & \textbf{.312} & .315 & .165 & \textbf{.167} & .170 & .054 & \textbf{.055} & .057 & 62.99 & \textbf{66.08} & 71.53 & .232 & \textbf{.234} & .239 \\
				\hline				
				\multicolumn{25}{l}{$^{\mathrm{a}}$ Test results from Mondays to Fridays (in sequential order) for weeks 32 and 33.}
			\end{tabular}
		}
		\label{tab:exp_split_412}
	\end{center}
\end{table*}

\begin{figure}
	\centerline{\includegraphics[width=0.48\textwidth]{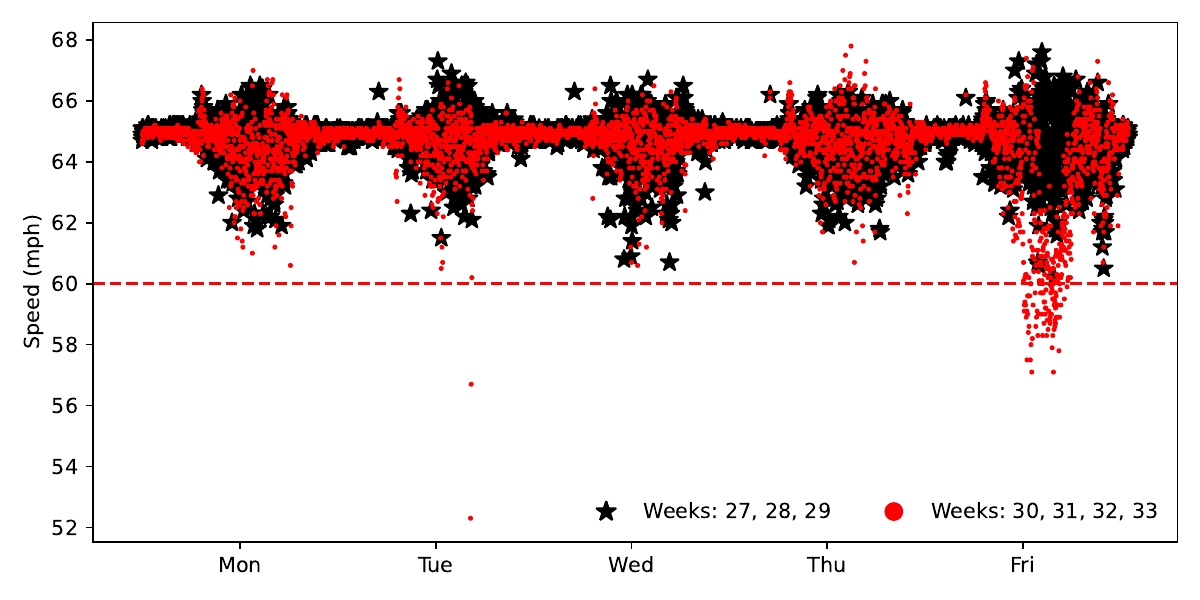}
	}
	\caption{Speed recorded by the PeMS 320287 loop detector in the segment  of the US50-E highway served by BS~320287 during weeks 27--33 in 2022. The red line shows the threshold applied to discretize the speed into speed levels.}
	\label{fig:speed}
\end{figure}

Figure~\ref{fig:speed}
illustrates the PeMS measured average speed
during the studied time period
and indicates the threshold that
divides the speed into two speed levels.
Recall that we discretize the speed 
as in~\cite{vesselinova2023road}.
Specifically, speeds above 60 mph 
correspond to a speed level 
different from those belonging to 
the 50--60~mph interval.
Notice that the speed level 
is constant during 
weeks 27--29
and hence, it does not
bring any information 
to the LSTM model
during training. 
In contrast, 
during the subsequent 30--33
weeks---on Wednesday (week 30) 
and on all Fridays---the speed level 
does change,
Figure~\ref{fig:speed}.
When examining the model's performance
on a per day level---Table~\ref{tab:mixture_split_322}
and Table~\ref{tab:exp_split_322}---the 
prediction errors are larger 
on Fridays but smaller on 
all other days 
when  contrasting 
the heterogeneous with 
the homogeneous data.
This is hinted by the MSE and RMSE 
results shown in Figure~\ref{fig:improvement_data_split}
as these measures penalize large errors,
and are sensitive to outliers.
The results reported
in Table~\ref{tab:mixture_split_322}
to Table~\ref{tab:exp_split_412}
are from 10 simulation runs.

In summary, when the model 
is trained with weeks 27--29 
of the heterogeneous data,
it is not exposed to 
varying speed levels
and hence, cannot learn the interrelation
between flow, speed and calls.
Thereby, the large prediction errors
evidenced on Fridays when the speed level 
does drop, Table~\ref{tab:mixture_split_322}
and Table~\ref{tab:exp_split_322}.
Contrarily, when the LSTM model
is trained with one more 
week---week 30 during which
 speed level alterations
are registered---the 
model learns the 
correlation between 
the three variables
and makes more accurate predictions
than by call time series alone,
Table~\ref{tab:mixture_split_412} and
Table~\ref{tab:exp_split_412}.

Overall, the results from this study 
highlight the relevance of data
when learning from data---a 
fundamental principle, 
sometimes neglected or traded for 
larger and complex learning structures.

\section{Discussion}
\label{sec:discuss}

\subsection{Novelty}

Our principal interest is 
in employing data 
that brings information about
the number of potential sources
of cellular load and their mobility. 
The fundamental premise is that 
substantial rather than 
incremental improvements 
in prediction accuracy
could be achieved with 
data describing 
the cellular load generation process.
In contrast to exogenous information 
(PoI and KG), 
such intrinsic data fosters sustainable solutions. 
The accuracy of 
the predictions---in relation to 
data---no longer dependents on 
external, static information
and its modeling but on 
how well the data represents
population dynamics.

\subsection{Impact}

During extreme events 
such as pandemics, 
human activity and mobility
together with the use of urban spaces
can drastically change~\cite{osorio2024analyzing, bouzaghrane2024human}.  
The use of the mobile network 
might intensify and that of 
mobile applications can diversify~\cite{zanella2022impact}. 
When containment measures
are in force, 
the divergence from
common patterns
could occur over 
a course of a day.
Then, information about
a BS operating in
a specific type of region 
becomes stale and 
not useful to the prediction task.
In other words,
the performance 
of models that rely on
exogenous information, 
depends on 
whether the external 
data represent
valid relationships 
between
the population 
and its use of 
the mobile system
in time and space
and falls short 
when the contextual information
is no longer valid.

Conversely, the power of 
learning the load generation process 
is in effectively dealing with 
any possible scenario.
Predictions are based on 
population dynamics, 
not on potentially 
outdated patterns
such as 
correlations between 
mobile traffic and 
urban layout,
which might  become irrelevant over time.

Employing population dynamics 
has the potential
not only to reduce the uncertainty 
about future cellular volumes 
but also to cut down 
the cost of 
collecting big and diverse data 
and hence, to decrease 
the computational complexity 
and memory requirements 
of the learning models.
Ultimately, 
this strategy could diminish 
the energy expenditure too.
Furthermore, concept drift \cite{dl2025ntp} 
arising from changes in network configurations 
or conditions and users’ behavior would have a
major impact on models relying on exogenous 
data (KG, PoI or other external
information) as these would need to be updated first.

\subsection{Implementation}

Modern transportation systems 
have monitoring capabilities---via 
loop detectors, sensors 
and video cameras---to
track road related metrics,
including flow and speed. 
Similar data can be 
collected by smart vehicles
and interchanged with
and send to RSU
or the mobile cellular system.
On highways, vehicular data 
would truthfully describe 
population dynamics. 
In cities, the monitored speed and flow 
can still be employed to forecast 
the cellular load 
placed by vehicles
and their passengers.
Dedicated network slices 
can then be managed efficiently 
to ensure time- and safety-critical applications
and smooth service
to fast moving users.

\section{Conclusion}
\label{sec:conclusion}

We study mobile traffic forecasting
from the perspective of data.
We porpose to empower 
the prediction model 
with information on 
the fluctuating number of mobile users.
To the best of our knowledge, 
this is the first approach that 
employs variables intrinsic to the 
mobile traffic generation process.
In a highway scenario, we employ 
vehicular flow and average speed 
as an estimate of 
the potential sources of
cellular load---vehicles and their passangers---and their mobility.
The prediction performance 
is largely and consistently improved---when 
contrasted with forecasting based on 
purely cellular time series---across 
diverse road traffic and mobile load conditions.
The approach does not rely on
collecting large volumes of exogenous information
but combines readily available data 
from the transport and mobile networks.
It has the potential to make 
mobile cellular forecasting 
more sustainable and accurate, 
especially when deployed on large-scale.

\bibliography{references}

\end{document}